\documentclass[journal]{IEEEtai}

\usepackage[colorlinks,urlcolor=blue,linkcolor=blue,citecolor=blue]{hyperref}

\usepackage{color,array}

\usepackage{graphicx}

\usepackage{graphicx,subfig,float}
\usepackage{subcaption}

\usepackage{subfloat}

\usepackage{amssymb}
\usepackage{algorithmic}
\usepackage{algorithm}
\usepackage{times}
\usepackage{latexsym}
\usepackage{amsmath}
\usepackage{booktabs}
\usepackage{array}
\usepackage{multirow}
\usepackage{diagbox}
\usepackage[most]{tcolorbox}
\newtcolorbox{mybox}[2][]{colback=gray!20,
    colframe=black,
    fonttitle=\bfseries,
    title=#2,
    #1}
\usepackage{colortbl}
\usepackage{wrapfig}
\usepackage{makecell}
\usepackage{hyperref}
\hypersetup{
    colorlinks=false,
    linkbordercolor=red,
    citecolor=red,
}
\usepackage{url}
\usepackage{subcaption}

\usepackage{threeparttable}
\usepackage{xcolor}

\newcommand{\sysname}{\textit{Inner-Probe}}

\definecolor{keywordcolor}{rgb}{0.0, 0.0, 0.0}

\newcommand{\PARAMETERS}{\item[\textbf{\color{keywordcolor}{Parameters:}}]}

\begin{document}

\title{Inner-Probe: Discovering Copyright-related Data Generation in LLM Architecture}

\author{Qichao Ma, Rui-Jie Zhu, Peiye Liu, Renye Yan, Fahong Zhang, Ling Liang, Meng Li, \IEEEmembership{Member, IEEE}, Zhaofei Yu, \IEEEmembership{Member, IEEE}, Zongwei Wang, \IEEEmembership{Member, IEEE}, Yimao Cai, \IEEEmembership{Member, IEEE}, Tiejun Huang, \IEEEmembership{Senior Member, IEEE}
\thanks{Manuscript received 26 February 2025; revised 2 October 2025; accepted 11 December 2025. (Corresponding author: Ling Liang.) }
\thanks{Qichao Ma and Tiejun Huang are with the 
School of Computer Science and the State Key Laboratory of Multimedia Information Processing, Peking University, Beijing 100871, China (e-mail: qcma@stu.pku.edu.cn, tjhuang@pku.edu.cn).}
\thanks{Zhaofei Yu is with the Institute for Artificial Intelligence, Peking University, Beijing 100871, China (e-mail: yuzf12@pku.edu.cn).}
\thanks{Rui-Jie Zhu is with the department of Electrical and Computer Engineering, University of California, Santa Cruz (e-mail: rzhu48@ucsc.edu).}
\thanks{Peiye Liu is with Alibaba DAMO Academy, Beijing 100102, China (e-mail: liupeiye.lpy@alibaba-inc.com).}
\thanks{Fahong Zhang is with College of Computing and Data Science, Nanyang Technological University, Singapore (e-mail: faar0220@gmail.com).}
\thanks{Renye Yan, Ling Liang, Meng Li, Zongwei Wang, and Yimao Cai are with the School of Integrated Circuits, Peking University, Beijing 100871, China. Zongwei Wang is also with the YanXin MicroElectronics Co., Ltd.(YXME), Shanghai, China (e-mail: victory@stu.pku.edu.cn, lingliang,meng.li,wangzongwei,caiyimao@pku.edu.cn).}
\thanks{This paragraph will include the Associate Editor who handled your paper.}}

\markboth{Journal of IEEE Transactions on Artificial Intelligence, Vol. 00, No. 0, Month 2020}
{First A. Author \MakeLowercase{\textit{et al.}}: Bare Demo of IEEEtai.cls for IEEE Journals of IEEE Transactions on Artificial Intelligence}

\maketitle

\begin{abstract}
Large Language Models (LLMs) utilize extensive knowledge databases and show powerful text generation ability. However, their reliance on high-quality copyrighted datasets raises concerns about copyright infringements in generated texts. Current research often employs prompt engineering or semantic classifiers to identify copyrighted content, but these approaches have two significant limitations: (1) Challenging to identify which specific sub-dataset (e.g., works from particular authors) influences an LLM's output. (2) Treating the entire training database as copyrighted, hence overlooking the inclusion of non-copyrighted training data.

We propose \sysname{}, a lightweight framework designed to evaluate the influence of copyrighted sub-datasets on LLM-generated texts. Unlike traditional methods relying solely on text, we discover that the results of multi-head attention (MHA) during LLM output generation provide more effective information. Thus, \sysname{} performs sub-dataset contribution analysis using a lightweight LSTM-based network trained on MHA results in a supervised manner. Harnessing such a prior, \sysname{} enables non-copyrighted text detection through a concatenated global projector trained with unsupervised contrastive learning. \sysname{} demonstrates 3× improved efficiency compared to semantic model training in sub-dataset contribution analysis on Books3, achieves 15.04\%–58.7\% higher accuracy over baselines on the Pile, and delivers a 0.104 increase in AUC for non-copyrighted data filtering.
\end{abstract}

\begin{IEEEImpStatement}
Large Language Models (LLMs) show remarkable text generation capabilities, but their reliance on high-quality copyright datasets has raised significant concerns about infringement. Existing methods include text matching, memorization-based detection, and prompt engineering. However, they fall short in capturing sufficient semantic information with extended contexts, and incur high computational costs. To overcome these limitations, we introduce a lightweight framework that leverages multi-head attention outputs for precise sub-dataset contribution analysis and effective non-copyrighted content filtering. Our approach achieves over 94.9\% accuracy in identifying the contributing copyrighted sub-datasets and 93.83\% accuracy with an AUC of 0.954 in filtering, substantially outperforming prior methods. This not only mitigates the shortcomings of prior research but also provides a practical, efficient solution for ensuring responsible real-world LLM deployment while protecting intellectual property.
\end{IEEEImpStatement}

\begin{IEEEkeywords}
Large Language Model, Copyright, Security.
\end{IEEEkeywords}

\section{Introduction}

Following the release of GPT-3 \cite{gpt-3}, LLM-driven applications have gained increasing attention in both industry and academia.  As transformer architectures mature, the demand for high-quality data has become urgent \cite{prove_data_quality}. However, a major security concern that hinders the establishment of the LLM community is the challenging task of protecting dataset copyright. Specifically, the current model providers lack the capability to identify which datasets contribute most to a response. This inability makes data holders hesitant to expose their datasets to LLM training since copyrighted datasets are of high quality but no credit is given to themselves. Data erasure is the direct method to stop LLM generating copyrighted texts, but the need for finetuning in LLM unlearning approach is computationally heavy and unreliable \cite{harry, arxivunlearn, unlearningdowngrade, detecting}. Conversely, if model providers could pinpoint the contributions of datasets to a response, a beneficial transactional framework between the data holder and LLM customer could be established \cite{Cui2024RethinkingUO}. 


 To establish a comprehensive framework for copyright dataset contribution analysis, two key functionalities are essential. First, the framework can evaluate the contribution of each copyrighted sub-dataset to responses generated by an LLM model. Second, it should filter out responses that do not present copyright concerns.

The contribution analysis of sub-datasets to LLM outputs can be framed as a data impact assessment problem, offering a broader application beyond copyrighted data.  Typically, mathematical, explainability-based attribution, and data traceability methods are developed for this task. Mathematical methods adopt a probabilistic approach to compute the similarity between texts and each sub-dataset, such as K-Nearest Access-Free (KNAF) \cite{follow, 2_privacy}. However, these studies mainly focus on theoretical proof based on strongly ideal assumptions with unknown parameters, which is hard to deploy in practice. Explainability-based attribution methods are separated into fine-tuning-based and prompt-based paradigms \cite{mezzi2025owns}. The former utilizes Shapley values (SHAP) and Local Interpretable Model-Agnostic Explanations (LIME), which explain output behaviors at the token or data sample level \cite{SHAPcite, LIMEcite}. However, it is challenging to be directly employed in LLMs due to the huge parameters and unrealistic gradient computations \cite{Anthropic, sharply}. Leveraging the reasoning capability of LLMs, the prompt-based paradigm aims to ask `what the response is based on', emphasizing factual grounding in cited documents rather than identifying styles or sub-dataset
contributions. This approach differs in goals and may amplify hallucinations \cite{hal}.  Data traceability refers to tracking the origin, process, and destination of data. Despite studies in large-scale commercial database management \cite{karkovskova2021design, tang2019sac}, currently, there is no investigation into the LLM generation process.

For non-copyright content filtering, prior studies have explored generated content identification, textual similarity, and LLM memorization \cite{mezzi2025owns}. Generated content identification includes watermarking and fingerprinting.  Data watermarking embeds labels into specific copyrighted data, facilitating insertion-based source tracing \cite{distribution, linguistic, entropy}. However, this approach is vulnerable to attacks, and the requirement of finetuning LLMs imposes huge computational overhead \cite{panaitescu2024can, training1, wasa}.  Fingerprinting passively extracts inherent characteristics from content to distinguish it from other similar objects, enabling identification of LLM-generated text. But it mostly refers to model-level copyrights, not copyrighted datasets \cite{llmmap}. Textual similarity approaches build copyright detectors based on lexical overlap, using metrics like LCS and Rouge-L \cite{liu-etal-2024-shield}, or contextual language models like BERT \cite{cho2022enhancing}. However, these methods primarily focus on textual content rather than the text generation process during LLM decoding. This limitation means they may miss cases where copyright-infringing texts depict entirely different scenarios while sharing high-level patterns that textual analysis alone cannot effectively capture. Lastly, previous studies have examined whether specific copyrighted data is included in training datasets \cite{decop, detecting},  using the memorization abilities of LLMs. However, these methods are limited to analyzing source datasets and fail to address copyright issues in LLM-generated outputs.

Based on the above analysis, previous studies always face three challenges in developing an efficient sub-dataset level copyright detection framework.
First, textual information alone is insufficient for accurate copyright detection, as it fails to capture higher-level semantic similarities such as character relationships or settings in a story. Second, current approaches separate sub-dataset contribution identification from non-copyright content filtering, resulting in fragmented methods that lack a comprehensive framework. Third, due to the large size and complexity of LLMs, the framework must be lightweight and efficient to enable real-time copyright analysis without imposing significant computational overhead.

To efficiently protect dataset copyright in LLMs, we propose \sysname{}, a lightweight framework that extracts inner features from LLM response to calculate the contribution of copyrighted sub-datasets and filters non-copyrighted LLM outputs. We begin by defining real-world copyright scenarios in LLM-based applications and identifying the most effective information for analyzing LLM behavior. Next, we develop an LSTM-based lightweight model for sub-dataset contribution analysis, integrating the response generation process during LLM decoding. Building on this LSTM model, we incorporate an efficient built-in detector to filter out non-copyrighted responses. Our contributions can be summarized as follows:

\begin{itemize}
\item We first define the copyright issues in real-world LLM applications, outlining the copyright identification workflow and the criteria for defining copyrighted sub-datasets. Subsequently, we emphasize that multi-head attention (MHA) layers offer efficient insights beyond semantic information and feed-forward network (FFN), supported by both empirical evidence and formal analysis.
\item We develop \sysname{}, a lightweight LSTM-based framework that integrates copyrighted sub-datasets contribution analysis and non-copyright content filtering functions. Specifically, the LSTM model in \sysname{} leverages multi-head attention (MHA) from both temporal and spatial perspectives during the LLM decoding stage. \sysname{} first trains the LSTM model in a supervised manner to enable the contribution analysis function. Building on the trained LSTM model, a global projection module is then trained using unsupervised contrastive learning to perform non-copyright content filtering.
\item We conduct extensive experiments to validate the effectiveness of \sysname{} in real-world applications. For the copyrighted sub-dataset contribution identification task, \sysname{} achieves an accuracy of over 94.9\% with a training time of less than 1.5 hours. For the non-copyrighted content filtering task, \sysname{} attains an accuracy of 93.83\% and an AUC score of 0.954 in the Pile dataset \cite{gao2020pile}.
\end{itemize}

\section{Background}

\subsection{Positioning and taxonomy} 
At present, there exist four main attribution method classes: (a) content identification (watermarking, fingerprinting), (b) security-based techniques (blockchains, digital signatures, zero-knowledge proofs, synthetic media forensics), (c) explainability methods (fine-tuning and prompt-based), and (d) data traceability (data lineage) \cite{mezzi2025owns}. 

Accordingly, we position \sysname{} within data traceability \cite{mezzi2025owns}. The first two classes answer whether specific copyrighted content is present. But they do not quantify how underlying training datasets contribute to a given generation. The latter two classes apply to LLM training data attribution. However, explainability methods often suffer from hallucination or require LLM fine-tuning costs to yield stable results. Meanwhile, existing data-traceability works concentrate on enterprise data management rather than tracing contributions through the LLM generation process \cite{karkovskova2021design, tang2019sac}. To contextualize our evaluation, we also draw on a survey of prompt-based LLM attribution. Following its taxonomy, we include a representative post-generation attribution (explainability) method as one of our baselines \cite{attribution_survey}.

\subsection{Limitations of current copyright detection}

\label{text_bf:copyright_takedown}
\textbf{Limitations of copyright detection on small models}

Before LLMs, techniques for copyright detection focus on learning-based text matching, which relies on small models. Text matching utilizes text similarity methods to match copyrighted texts for those fragments. Statistical one calculates word frequency using TF-IDF and Simhash \cite{jin2017research}. However, it ignores text semantics, and is limited to pair-wise comparison, requiring separate computations for each document pair. Deep learning ones use SVM with hand-crafted features \cite{prieto2018supervised}, autoencoder \cite{briciu2021autoat} and bi-directional LSTM (BiLSTM) \cite{tavan2021bert, petropoulos2024roberta} to capture deep semantic information. Whereas, such methods are only tailored for specific text types and are prone to overfitting, such as newswire stories \cite{briciu2021autoat} or comics \cite{xin2024text}. More importantly, the definition of copyright infringement is limited to simple text manipulations, including random crop and back-translation of original copyrighted texts \cite{xin2024text}. This fails to align with copyright legal definitions, and is inadequate for LLM-generated texts, which has complex variations and adaptations of original copyrighted works.

\textbf{Limitations of existing methods on LLM copyright detection}
Currently, most LLM copyright detection frameworks focus on identifying whether a text exists in the entire training database \cite{detecting, decop}. These methods typically utilize the memorization capabilities of LLMs, which can overemphasize sub-datasets that do not raise copyright issues. This results in an incomplete approach, as a comprehensive real-world copyright analysis framework has not yet been fully defined. Additionally, existing studies are constrained by limited context lengths (typically up to 200 tokens) and focus primarily on datasets like books and news \cite{copyrighttakedown}. Thus, there is a need for a more concrete LLM copyright framework and a broader testing environment that encompasses diverse content and longer context lengths.

\subsection{Copyright Protection}

\textbf{LLM unlearning} 

Research shows that LLMs can memorize training datasets, leading to efforts to make LLMs forget the data. This method is categorized into parameter-tuning and parameter-agnostic ones. Parameter-tuning involves fine-tuning such as data sharding \cite{liu2024forgetting}, gradient-based finetuning with reverse loss \cite{yu2023unlearning}, knowledge distillation \cite{harry}, additional learnable layer \cite{chen-yang-2023-unlearn}, or parameter-efficient module operation \cite{hu2024separate}. However, these methods can inadvertently affect unrelated information due to knowledge conflicts. To alleviate such fine-tuning costs, the parameter-agnostic method utilizes in-context learning, but the effect is limited to single conversations \cite{DBLP:conf/icml/PawelczykNL24}. Most importantly, using a Min-k probability attack, the unlearning methods prove unreliable \cite{detecting}. 

\textbf{Malicious inquiry protection}

Current studies attempt to prevent LLMs from generating copyrighted content by simple keyword detection and lexical metrics. For example, modifying LLM logits to avoid generating copyright contents by an N-Gram
model during decoding \cite{ippolito-etal-2023-preventing}. However, these methods remain vulnerable to malicious queries, such as the jailbreak attack that aims to bypass LLM safety checks. White-box attacks use adversarial approaches, using gradient \cite{zou2023universal}, logits \cite{zhou2024don}, or fine-tuning \cite{iclr24oral}, while black-box attacks use template completion \cite{deng2024pandora}, rewriting \cite{YuanJW0H0T24} and use LLM to generate attacks \cite{casper2023explore, DBLP:conf/camlis/BirchHTSG23}. The challenge arises because LLMs function as knowledge databases, making it more likely that copyrighted data is incorporated into generated texts without being directly quoted \cite{cyphert2023generative}.

\subsection{Architecture Basic in LLMs}

Most existing studies focus on textual information from the customer or data provider's perspective when addressing copyright issues in LLMs, with few considering the model provider's perspective. This oversight neglects the critical role of the generation process within the LLM architecture, which is essential for understanding how copyrighted content can emerge during text generation.

The architecture of LLMs consists of multi-head attention (MHA) and feed-forward network (FFN). In the MHA component, the attention scores are computed, allowing the model to focus on different segments across separate representation subspaces. The FFN then performs additional feature extraction, refining the information for further processing within the network.

For MHA, let $X, Y \in \mathbb{R}^{n \times d}$ denote the input and output of an MHA layer, where $X$ and $Y$ represent the embeddings of all tokens. The projection weights for the query, key, and value are represented by $W^Q, W^K, W^V \in \mathbb{R}^{d \times d_h}$, where $d$ represents the maximum token length supported by an LLM and $d_h$ is the dimension of the attention head. The MHA processes can be summarized as follows:
\begin{equation}
\small
    Q = XW^q, K = XW^k, V = XW^v, 
\end{equation}
\begin{equation}
\small
    Y =  softmax(\frac{Q K^T}{\sqrt{d}} )V
    \label{eq:dim}
\end{equation}
The FFN typically consists of two or three fully connected layers. It takes $Y$ as input, and its output $X$ serves as input to the next MHA.
\section{Copyrights in LLM}

Since current settings for copyright detection in LLM-based applications are inadequate, we first define the essential functionalities that a comprehensive copyright detection framework should include. Next, we focus on how to define copyrighted data within the context of LLM-based applications. Finally, we identify which intermediate features during the LLM decoding process provide the most valuable information for effective copyright detection.

\subsection{LLM-based copyright detection framework} 

\begin{figure}[htbp]
    \centering
    \includegraphics[width=0.98\linewidth]{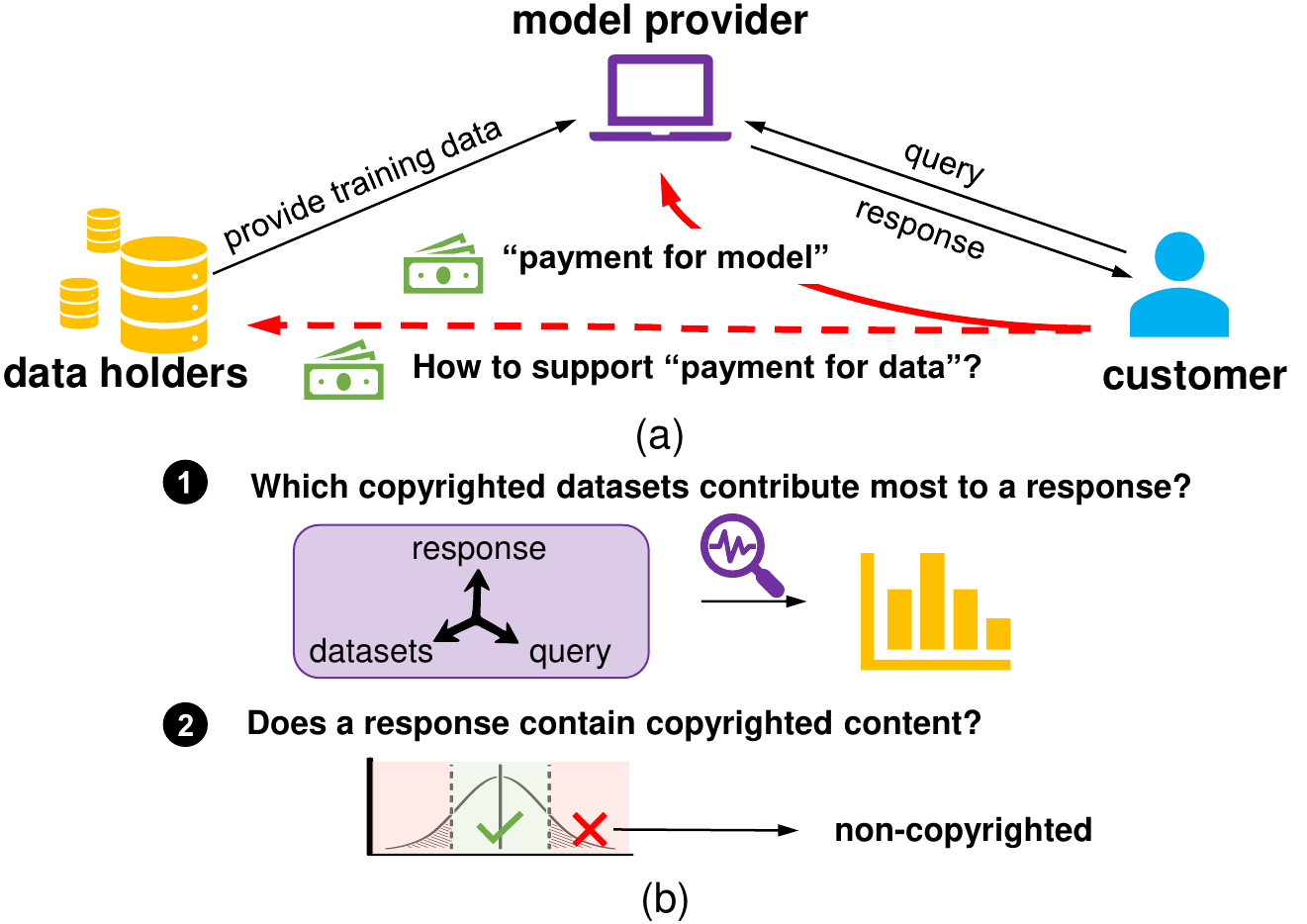}

	\caption{(a) LLM-based community and commercial behaviors; (b) both copyright detection and dataset contribution analysis are key to supporting `payment for data' in this transactional framework. }
	\label{fig:moti}
\end{figure}

In the LLM-based application shown in Fig. \ref{fig:moti}(a), three participants are involved: a customer, a model provider, and multiple data holders. Initially, data holders supply the training sub-datasets to the model provider to train a multifunctional LLM. Then, the customer can query the model provider and get the corresponding LLM response. Therefore, the customer should pay for the service once receiving the response. However, unlike a general point-to-point transaction, the customer should not only pay for the model provider but also pay for the data holders. The reason is that both the LLM and training datasets contribute to the responses received by the customer. The model provider typically charges the customer for various services, but the challenge of compensating data holders remains unaddressed. Therefore, a copyright detection framework is needed to bridge this gap.

Based on the above setting, the copyright detection framework, as shown in Fig. \ref{fig:moti}(b), must support two essential functionalities. First, the framework needs to efficiently analyze the contributions of each copyrighted sub-dataset to a given response produced by the LLM. Note that this contribution analysis can be applied to any response. However, not all responses will necessarily involve copyrighted material—some may rely only minimally on the copyrighted datasets, or not at all. As a result, the framework must also include a filtering mechanism capable of excluding responses that do not present any copyright concerns. This dual functionality ensures that the framework is both precise in identifying copyright-relevant contributions and efficient in excluding non-issue responses from further scrutiny.

\subsection{Copyrighted sub-datasets for LLM}

\begin{figure}[htbp]
    \centering
    \includegraphics[width=\linewidth]{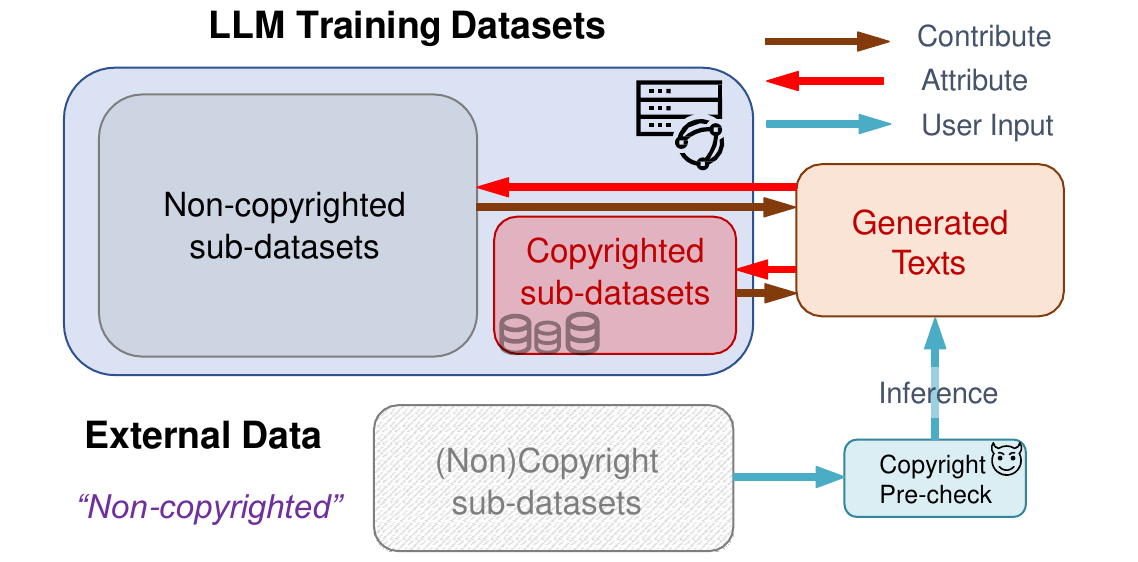}

	\caption{The range of data used in LLM training and inference.
 Only copyrighted sub-datasets used in LLM training are considered for contributing to LLM-generated texts. External copyrighted data can be pre-filtered during inference.}
	\label{fig:non-copyright}
\end{figure}

Unlike previous studies that treat the entire pre-training database as copyrighted, we define a more concrete copyright identification method suited to real-world scenarios. As shown in Fig. \ref{fig:non-copyright}, we focus on the impact of copyrighted sub-datasets in the LLM's training data. These high-quality sub-datasets are often more heavily relied upon during training, making it crucial to identify potential copyright issues within them. Non-copyrighted datasets include open-sourced data within the training set and external data used during the LLM inference phase.

\subsection{Problem formulation for LLM-based copyright detection}

Based on the settings for copyrighted sub-datasets in training data, we can formally define the two key functionalities that a copyright framework should support. Specifically, we consider the following scenario: given a user-defined text input $T$, an LLM $M$ generates a response $R$. The LLM model is trained on the entire dataset $D$ which includes copyrighted sub-datasets $D_c$ and non-copyrighted sub-datasets $D_{\bar c}$ that satisfy
\begin{equation}
\begin{gathered}
    D_{\mathrm{c}} = D \setminus D_{\mathrm{\bar c}}, \\
    D_{\mathrm{c}} = \bigcup_{i=1}^n D_{\mathrm{c}}^i, \quad D_{\mathrm{\bar c}} = \bigcup_{j=1}^m D_{\mathrm{\bar c}}^j.
\end{gathered}
\label{eq:composition}
\end{equation}
We use $D_c^i$ to represent the $i^{th}$ copyrighted sub-datasets, and similarly, $D_{\bar c}^j$ to represent the $j^{th}$ non-copyrighted sub-dataset.

\textbf{Copyrighted sub-datasets contribution analysis} For a generated result $R$, we use a soft contribution score $S \in \mathbb{R}^n$ to quantify the relative influence of $n$ copyrighted sub-datasets. To compute this score, a parameterized mapping function  $F_{\theta}: { R\mapsto S} $ is needed. However, directly solving this mapping function is challenging, as user responses may either exhibit uncontrolled behavior or include additional directional prompts that induce the LLM to generate copyrighted content, potentially disrupting its behavior. Furthermore, we observe that copyright infringements in generated texts can only originate from the training datasets. Therefore, we propose adopting a substitute approach to train the contribution analysis model based on the context within the original copyrighted sub-datasets. Specifically, we would like to optimize the mapping stragtegy  $F_{\theta'}: { T_k\mapsto S'_{k}} $, where $T_k \in D_c^i$. The objective function can be summarized as 
\begin{equation}
\min _{\theta'} \mathcal{L}(\theta') = \min _{\theta'}\left(-\sum_{k=1}^{|D_{\mathrm{c}}^i|} \log \left(F_{\theta'}\left(T_k\right), h_i\right)\right),
\label{eq:sim_eq}
\end{equation}

$h_i$ is a one-hot vector indicating the index of a copyrighted sub-dataset. Next, we can use the induced response $R$ to evaluate the performance of the trained mapping function $F_{\theta'}$. Theoretically, we have 
\begin{equation}
\label{eq:6}
F_{\theta'}: { R \mapsto  S'},  F_{\theta}: { R \mapsto  S}.
\end{equation}
In  Eq. \ref{eq:6}, once $R$ is generated through a deliberately designed induced prompt, the actual distribution of $S$ can be obtained. The performance of the trained mapping function $F_{\theta'}$ can then be assessed using metrics such as the MSE or KL divergence between $S'$ and $S$.

\textbf{Filtering out non-copyrighted responses}
To achieve the goal of non-copyrighted response filtering, the key is to build a confidence score function $Conf()$ that takes model $M$ and response $R$ as input to evaluate how closely $R$ aligns with the copyrighted sub-datasets $D_c$. This can be formulated as 
\begin{equation}
I_\delta({M, R})= \begin{cases}  R \in \text{Copyrighted} &  Conf({M, R}) \geq \delta, \\  R \in \text{Non-copyrighted}  &  Conf({M, R})<\delta.\end{cases}
\label{eq:decision}
\end{equation}

Here $\delta$ represents the filtering threshold, and the filtering decision is returned by the function $I_{\delta}()$.

\subsection{Analyzing LLM architecture: Hidden States for Causality}

\label{sec:mha}

To achieve copyright identification for an LLM, it is essential to determine what information should be utilized during the inference stage. In this subsection, we first analyze the impact of datasets on hidden state representations by comparing models trained and untrained on specific input data. We then demonstrate that MHA exhibits stronger causality compared to FFN.

To investigate the architectural information of LLMs beyond textual characteristics, we adopt UMAP, a popular dimensionality reduction method that uses graph layout algorithms to map data into a low-dimensional space \cite{mcinnes2018umap}. The experiment uses two text sets: (1) 8 classes of original texts from the Pile \cite{gao2020pile}, and (2) corresponding LLM-generated texts through `continuation' prompts by taking original texts as input, simulating a copyright-related generation scenario while maintaining the same class.

\textbf{\textit{Observation 1: When processing inputs, an LLM's hidden states (MHA, FFN) encode richer information for input data that appears in its training dataset, compared to data not in it. This phenomenon persists in two scenarios: (1) when directly inputting original training data, and more importantly (2) when inputting text generated by the LLM using prompts derived from training data.}}

\begin{figure}[t]
    \centering
    \includegraphics[width=\linewidth]{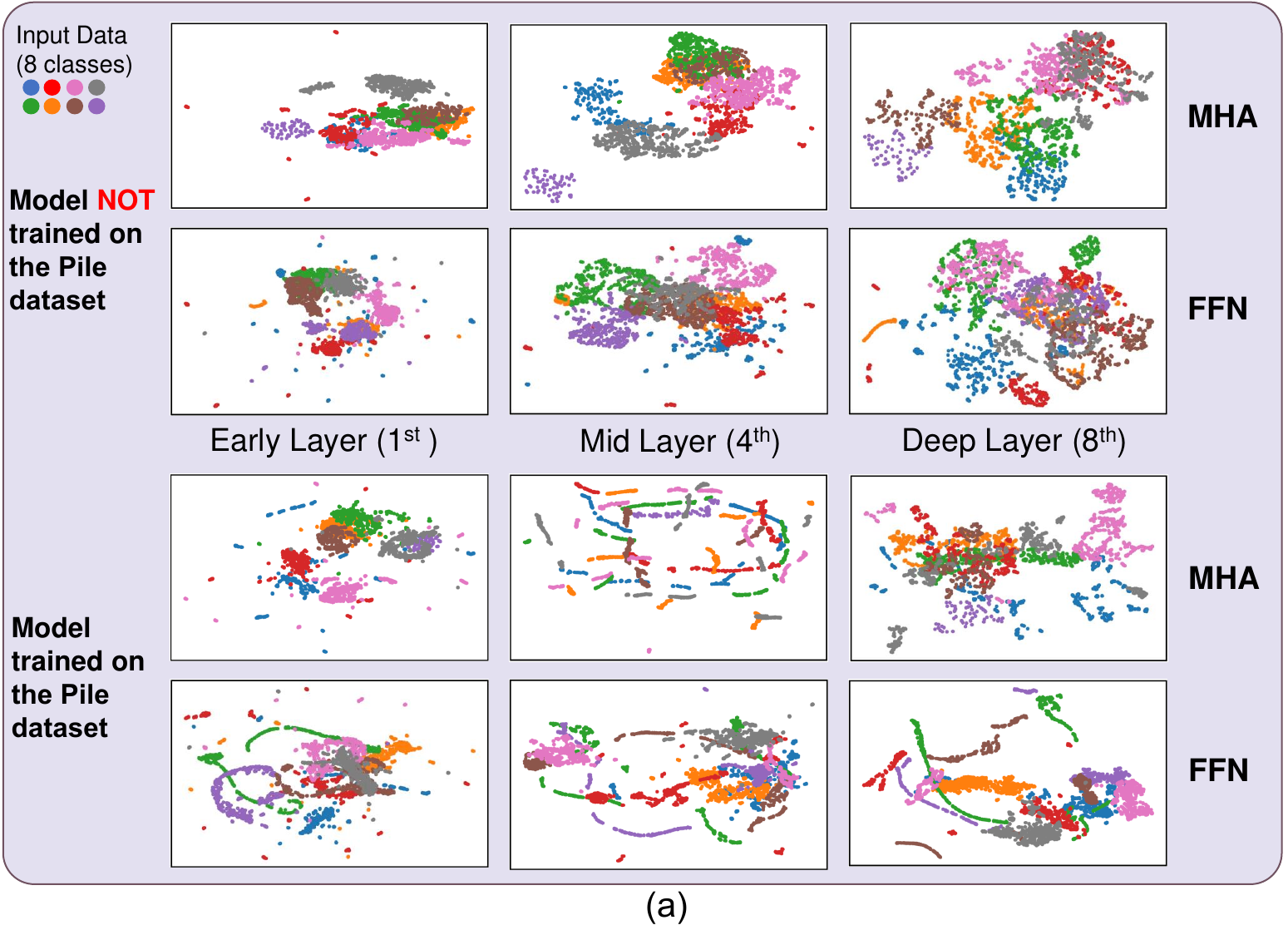}

    \vspace{0.1cm}
    \includegraphics[width=\linewidth]{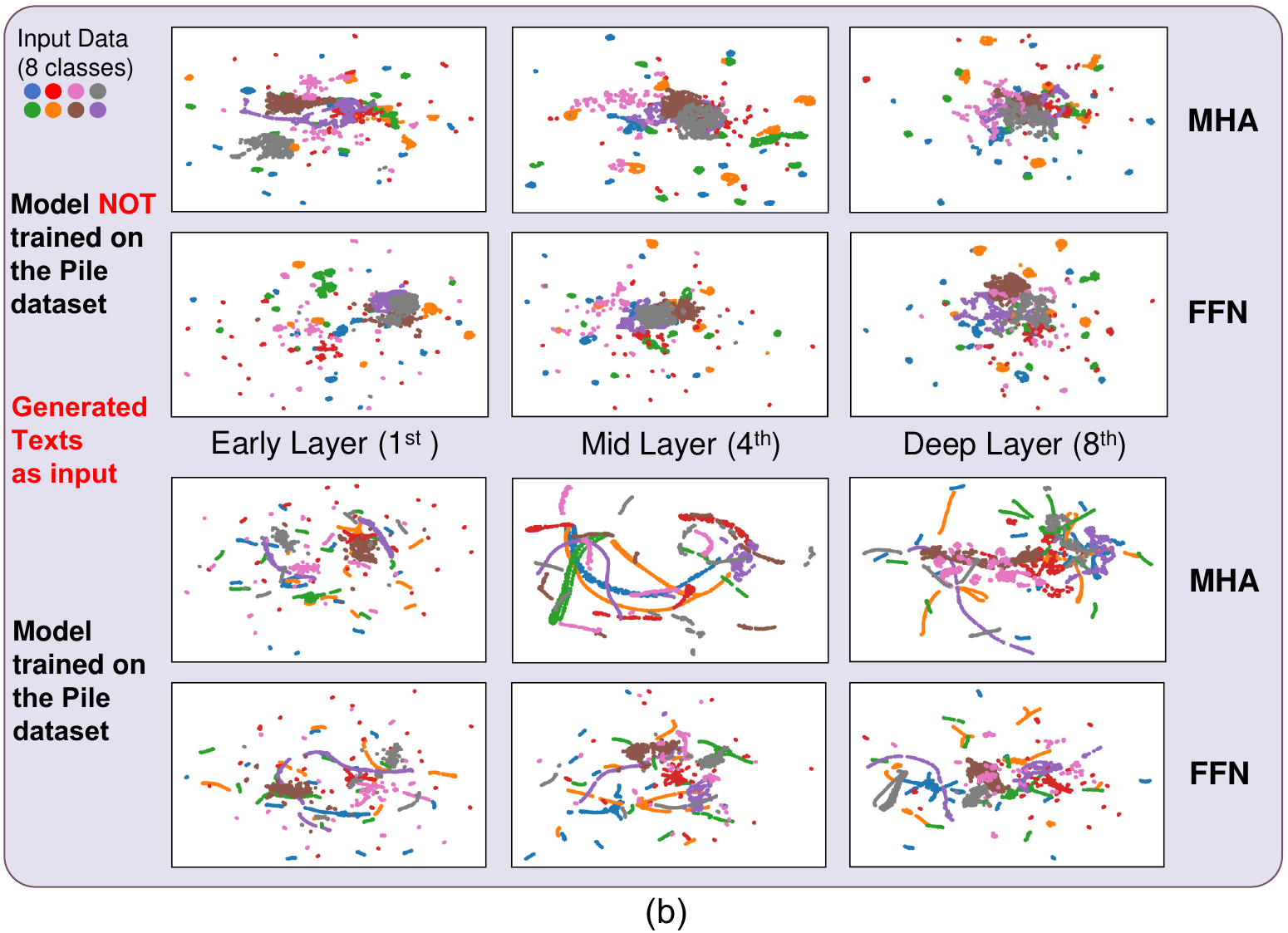}

    \caption{Visualization of the statistical differences (UMAP) in hidden states (MHA, FFN)  \textbf{(a)} with 8-class input texts from data in the Pile; \textbf{(b)} with input texts that are generated from the pile  \cite{gao2020pile}. For each subfigure in both (a) and (b),  \textbf{Top Row}: MHA/FFN output visualization from an LLM (BERT) not trained on the Pile. \textbf{Bottom Row}: MHA/FFN output visualization from an LLM (GPT-series) trained on the Pile across layers.}
    \label{fig:umap_vis}
\end{figure}

The results in Fig. \ref{fig:umap_vis}(a) reveal that models not trained on the Pile only show clustering behavior in layers 1,4 and 8, indicating that no additional information can be directly found among these layers. In contrast, for models trained on the Pile, an additional trajectory behavior, i.e. MHA in layer 4, are observed in MHA and FFN layers.

Next, we extend experiments to generated texts, as shown in Fig. \ref{fig:umap_vis}(b). Compared with top 2 rows in Fig. \ref{fig:umap_vis}(a), clustering becomes vague in the same rows in Fig. \ref{fig:umap_vis}(b). This means for models not trained on the Pile, semantic features are inadequate for copyright analysis on generated texts. In contrast, in the bottom 2 rows where the model is trained on the Pile, trajectory patterns in Fig. \ref{fig:umap_vis}(a) are maintained in Fig. \ref{fig:umap_vis}(b).

\textbf{\textit{Observation 2: MHA expresses more causality than FFN, making it more suitable for copyright analysis.}}

\begin{figure}[htbp]
    \centering
    \includegraphics[width=\linewidth]{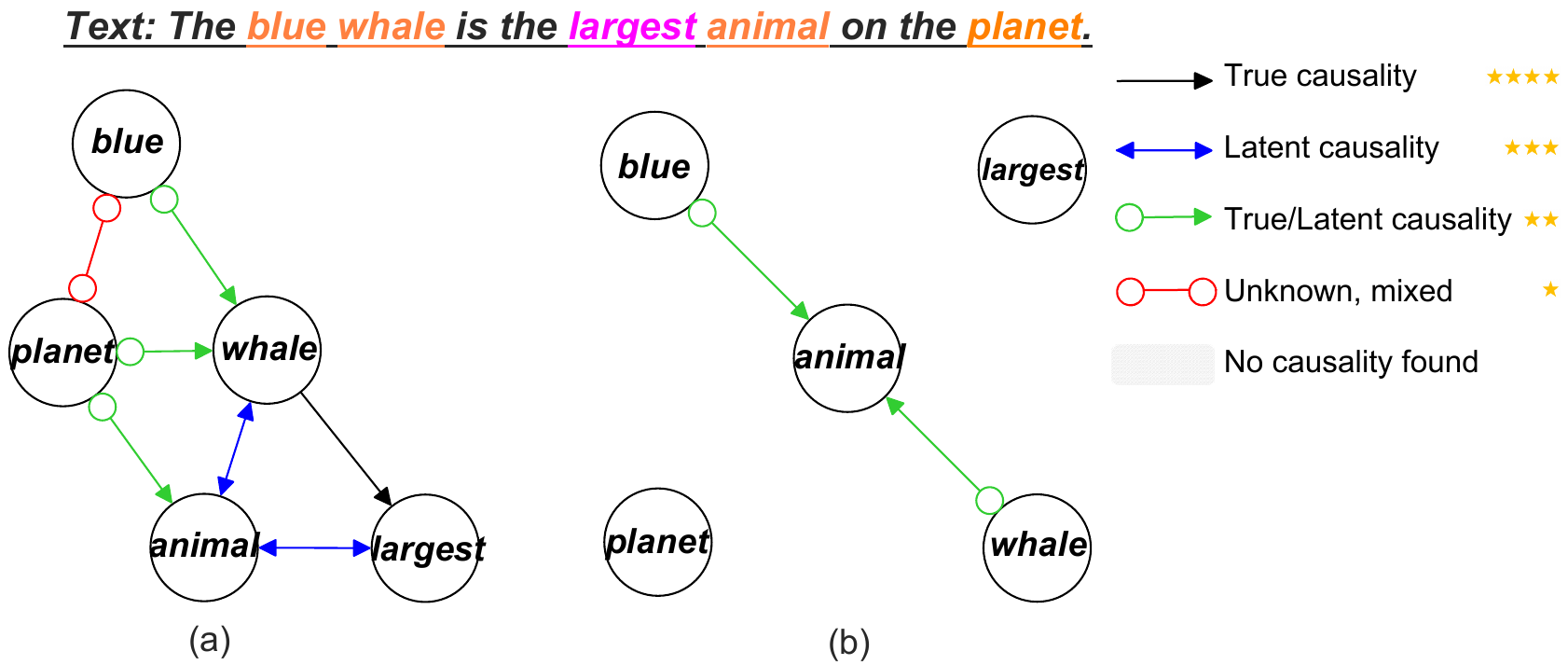}

	\caption{Comparison of a causal graph learned from MHA and FFN. The text is ‘The blue whale is the largest animal on the planet.' The graph learns causality comprehensively from MHA, but far less from FFN, (a) MHA result; (b) FFN result. }
	\label{fig:causal_graph_compare}
\end{figure}

We mainly focus on causal graphs to determine which hidden state benefits copyright protection. For variables $P$, if $P_j$ changes with $P_i$, then $P_i$ is considered the cause of $P_j$. Therefore, they construct a Partial Ancestral
Graph (PAG) where each parent node directly or conditionally causes its child nodes. This PAG can be considered a causal graph, with a complete causality list shown in Fig. 
\ref{fig:causal_graph_compare}.

To learn such a graph from hidden states in LLMs, probabilistic methods are needed. We utilize the Structured Causal Model (SCM) in \cite{neurips23causal}. We chunk MHA and FFN outputs into matrices, then two causal graphs learn from them. An example result is shown in Fig. \ref{fig:causal_graph_compare}
. We set the target token (marked in purple) and potential contributing tokens (marked in orange), which are analyzed during learning. We find that MHA constructs comprehensive relationships across tokens, but FFN captures only three token connections, which means MHA shows superior causality analysis for input texts.

We further support such findings by analyzing the covariance of MHA and FFN. Such covariance and its inverse directly reflect conditional dependencies, where non-zero off-diagonal elements mean dependencies between variables, and zero values show independence. Let the FFN output be $Y_2 = W_2\sigma(W_1Y + b_1) + b_2$, with $A = \operatorname{softmax}\left(\frac{Q K^T}{\sqrt{d}}\right)$. As in Eq. \ref{eq:nips23}, FFN output $Y_2$ applies the same position-invariant linear transform $W_1, W_2$ for each token index $i$. Thus, the covariance values are suppressed to near zero for $i \neq j$.

\begin{equation}
\begin{aligned}
C_{Y}=A C_{V} A^{\top}, \quad C_{Y_2}=W_2 C_{\sigma(W_1Y + b_1)} W_2^{\top}
\end{aligned}
\label{eq:nips23}
\end{equation}

\begin{figure}[htbp]
    \centering
    \includegraphics[width=\linewidth]{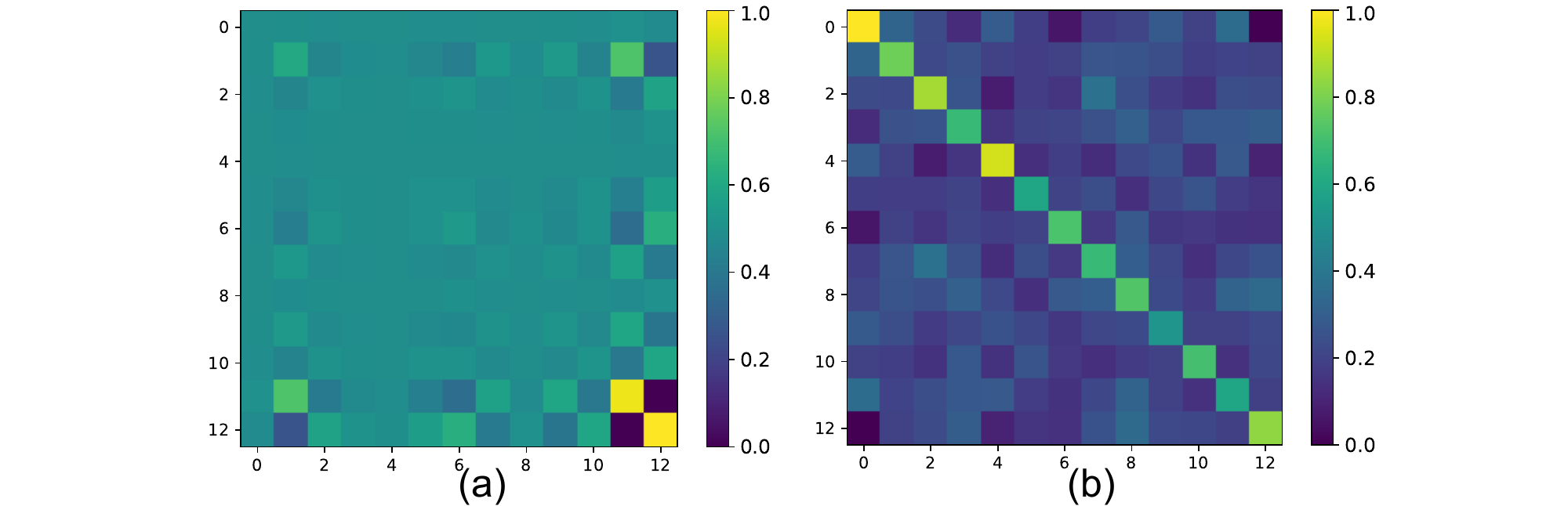}

	\caption{Comparison of inverse covariance matrix from MHA and FFN. The complete independence (diagonal) blocks FFN from modeling causality, (a) MHA result; (b) FFN result.}
	\label{fig:inv_cor}
\end{figure}

As verified in Fig. \ref{fig:inv_cor}, MHA flexibly captures token-wise dependencies. In contrast, FFN shows an almost completely diagonal inverse covariance matrix, blocking it from learning more causality.

\section{\sysname{} Framework}

\begin{figure*}[t]
    \centering
    \includegraphics[width=\linewidth]{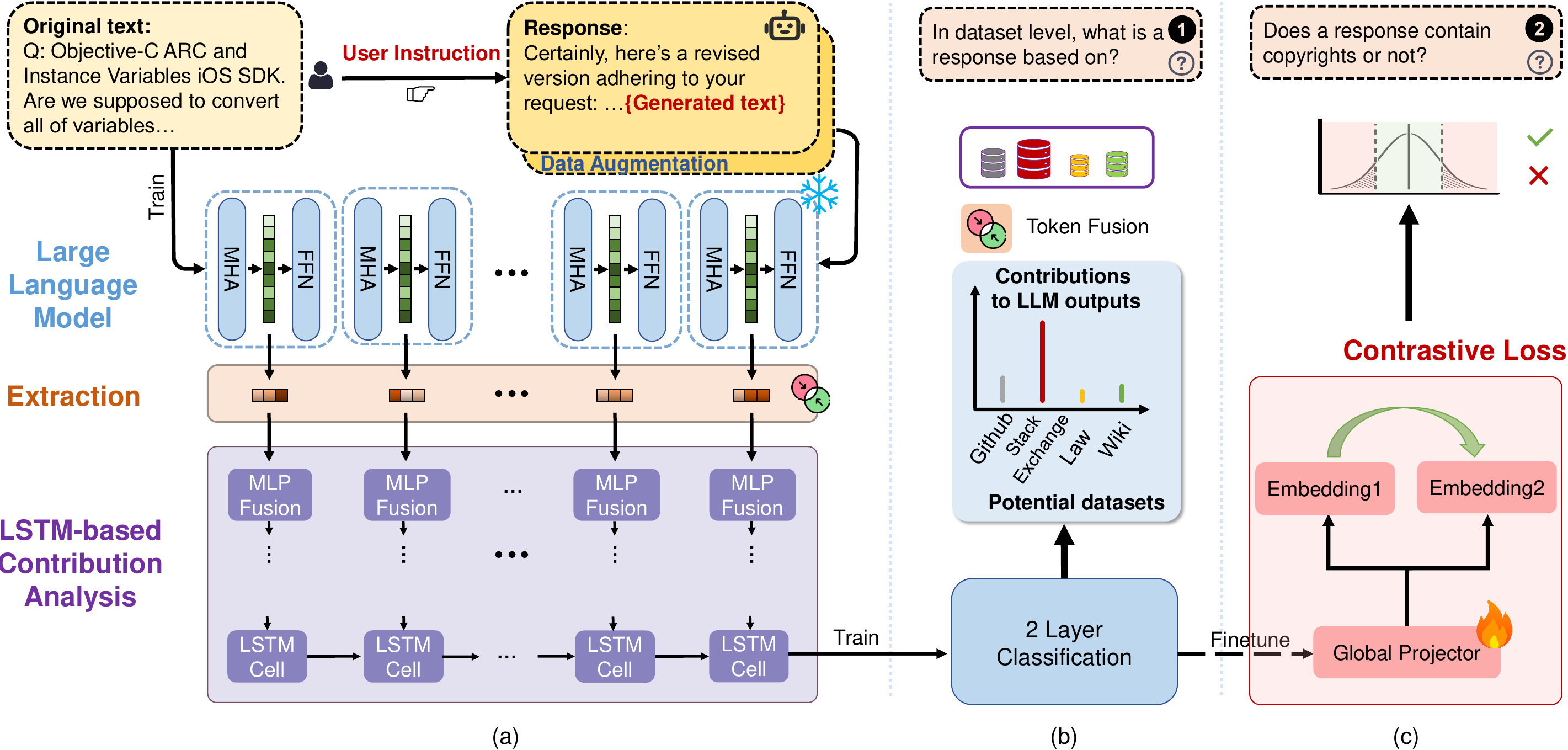}
    \caption{\sysname{} framework: (a) LSTM-based feature extractor module; (b) copyrighted sub-datasets contribution analysis module; (c) non-copyrighted responses filtering module.
    }
    \label{fig:whol}
\end{figure*} 

\subsection{Overview}

To address the copyright issue in LLM-based applications, we propose \sysname{}, a lightweight framework that integrates both copyrighted sub-dataset contribution analysis and non-copyright response filtering functionalities. The overall framework is illustrated in Fig. \ref{fig:whol}. After a response is generated by the LLM, we perform an additional prefilling process to extract MHA from the LLM.  The core of \sysname{} is built upon an LSTM model that takes the multi-head attention (MHA) from the LLM during prefilling process, as shown in Fig. \ref{fig:whol}(a). Once high-level information is extracted, the contribution analysis and response filtering are carried out by the respective adaptors, as shown in Fig. \ref{fig:whol}(b) and (c). 

During the \sysname{} training phase, the contribution analysis module is first trained in a supervised manner according to Eq. \ref{eq:sim_eq}. In this step, in addition to the trained contribution analysis adaptor, we also obtain an LSTM model that has learned the high-level representations of copyrighted sub-datasets. Next, \sysname{} freezes the LSTM model and uses it as a high-level feature extractor to train the non-copyright filtering adaptor through an unsupervised contrastive learning strategy.

\subsection{Copyrighted sub-datasets contribution analysis module}

In Sec. \ref{sec:mha}, we demonstrated that MHA contains significantly more effective information during the text generation process of an LLM model. However, the sequence length of an LLM response is dynamic, making it crucial to design a unified contribution analysis module that can handle varying sequence lengths. In this subsection, we propose two strategies for extracting fixed-length information from MHA with varying lengths, followed by theoretical explanations on how to assess the efficiency of information extraction.

\subsubsection{Information extraction and LSTM-based contribution analysis module}
~

The attention is sparse during LLM inference, where not all tokens carry effective information for an MHA layer. Based on this, we design two primary token extraction strategies: interval-based and variance-based, as shown in Alg. \ref{alg:all_token_selection}. These strategies transform the dynamic MHA lengths into a fixed size, making them more suitable for a copyright identification framework to handle various context lengths effectively.

\textbf{$k$ token interval sampling }(\textbf{\textsc{Inter}})

In this token extraction method, we directly sample $k$ tokens with interval $n/(k-1)$, where $n$ stands for the total number of tokens in LLM-generated texts, and $Y_l$ means MHA outputs in layer $l$. Specifically, when $k=1$, the middle token is sampled. The procedure can be summarized in Eq. \ref{eq:inter}. 
\begin{equation}
    \overline{O}_l=
    \begin{cases}
    \{Y_l[\frac{n}{2}]\} & k=1\\
    
    \{Y_l[i] ~|~ i=0,\frac{n}{k-1}, \frac{2n}{k-1}...\} & k > 1
    \end{cases}
    \label{eq:inter}
\end{equation}
Here, we denote $\overline{O}_l$ as the selected token representations in the $l^{th}$ MHA block. This sampling method is intuitive but effective at handling long-range context, which is denoted as Interval (Inter) for simplicity.

\textbf{Top-k pivotal tokens (\textsc{Var}, \textsc{A-Var})}

We define pivotal tokens as those with higher representation variance, which shows significance in lightweight LLM inference \cite{DBLP:conf/aaai/LeeJKKP24}. As shown in Eq. \ref{eq:var_sel}, for layer $l$, we calculate variance $Var(Y_{l}[i])$ for each token position $i$. The ones with top-k highest variance are selected, constructing extracted representation $O_l$.  As shown in Eq. \ref{eq:var_sel}, this method is annotated as Variance (Var).

\begin{equation}
    \overline{O}_l = \{ Y_{l}[i] \mid 0 \leq i < n,\, i \in \text{top-}k(\text{Var}(Y_{l}[i])) \}
\label{eq:var_sel}
\end{equation}

Furthermore, we propose Aligned-Variance (\textsc{A-Var}) strategy to identify consistent pivotal tokens across layers, as shown in Alg. \ref{alg:all_token_selection}. After selecting top-k token indexes from layer $l$ in \textsc{Var}, we aggregate the layer selections by counting how frequently each position appears in the top-k set across all layers. Specifically, for each position $i$, $Num_i$ accumulates its total occurrence count in all layers. We then select $k$ positions with the highest occurrence counts as $A_{index}$, which serves as a unified set of token positions for all layers. To sum up, the above two variance-based methods are designed for different granularities, with the first offering an understanding of specific layers and the second ensuring consistency and robustness. 

\begin{algorithm}
\caption{Information Extraction Strategies}
\label{alg:all_token_selection}
\small
\begin{algorithmic}[1]
\REQUIRE MHA Outputs (single head) $Y \in \mathbb{R}^{L \times n \times d}$ with $n$ tokens, $L$ layers, and $d$ dimensions
\PARAMETERS Strategy $s \in \{\textsc{INTER}, \textsc{VAR}, \textsc{A-VAR}\}$, Number of selected tokens $k$ per layer
\ENSURE A Set of Token Positions $A_{\text{index}} \subseteq \{0, \dots, n-1\}$, where $\mid A_{\text{index}} \mid = k$, and final extracted representation $O \in \mathbb{R}^{L \times k \times d}$

\IF{$s = \textsc{INTER}$}
    \STATE \textit{// Strategy 1: Interval Sampling}
    \IF{$k = 1$}
        \STATE $A_{\text{index}} \gets \{\lfloor n/2 \rfloor\}$
    \ELSE
        \STATE $\delta \gets \lfloor \frac{n}{k-1} \rfloor$
        \STATE $A_{\text{index}} \gets \{i\delta \mid i \in \{0,1,\dots,k-1\}\}$
    \ENDIF
\ELSIF{$s = \textsc{VAR}$}
    \STATE \textit{// Strategy 2: Layer-wise Top-K Variance}
    \STATE Initialize $A_{\text{index}} \gets \emptyset$
    \FOR{each layer $l \in \{1, \dots, L\}$}
        \STATE $v_i^l = \text{VAR}(Y_l[i])$
        \STATE $A_{\text{index}}^l \gets \text{TopK}(\{v_i^l\}_{i=0}^{n-1}, k)$
    \ENDFOR
\ELSE
    \STATE \textit{// Strategy 3: Aligned Variance (\textsc{A-VAR})}
    \STATE Initialize $A_{\text{index}} \gets \emptyset$
    \STATE Initialize $\text{Num}_i \gets 0, \forall i \in \{0, \dots, n-1\}$
    \FOR{each layer $l \in \{1, \dots, L\}$}
        \STATE $v_i^l = \text{VAR}(Y_l[i])$
        \STATE $\text{tmp}_l \gets \text{TopK}(\{v_i^l\}_{i=0}^{n-1}, k)$
        \FOR{each index $i \in \text{tmp}_l$}
            \STATE $\text{Num}_i \gets \text{Num}_i + 1$
        \ENDFOR
    \ENDFOR
    \STATE $A_{\text{index}} \gets \text{TopK}(\{\text{Num}_i\}_{i=0}^{n-1}, k)$
\ENDIF

\FOR{each layer $l \in \{1, \dots, L\}$}
    \STATE $O_l \gets Y_{l}[A_{\text{index}}^l]$
\ENDFOR

\RETURN $A_{\text{index}}, O$
\end{algorithmic}
\end{algorithm}

\textbf{LSTM-based Contribution Analysis}

The contribution analysis module and the LSTM are trained simultaneously in a supervised manner. During training, contexts from copyrighted sub-datasets are fed into LLM for prefilling, allowing the LSTM framework to learn layer-wise dependencies. This process is significant because reintroducing data from the LLM's training dataset into the model triggers layer-wise patterns, as highlighted in \textit{Observation 1}.

As depicted in Fig. \ref{fig:whol}(a),  the classifier captures the layer-wise relationship between extracted MHA outputs, inspired by \textit{Observation 1} in section \ref{sec:mha}. Therefore, the number of time steps in LSTM matches the layer number in LLM architecture. For each time step, the MLP fusion module first compresses $O_l$ into a compact representation, which is then fed into the corresponding LSTM cells. The final output of the LSTM is passed to a two-layer classifier, producing a softmax distribution that represents the contribution score of each copyrighted sub-dataset.

\subsubsection{Theoretical Explanations for Different Information Extraction Strategies}

Since LLM responses are uncontrolled and influenced by diverse user inputs, their evaluation may not accurately reflect the effectiveness of information extraction methods. How can we predict the performance of information extraction strategies for copyright contribution analysis and assess them comprehensively? 
To answer this, we propose a theoretical framework based on the Information Bottleneck.
Specifically, we use Mutual Information to measure how much task-relevant information is preserved after extraction.

Mutual Information (MI) quantifies the statistical dependence between two variables and the information they share. Formally, MI is defined in Eq. \ref{eq:MI}, where $p(a,b)$ is the joint probability distribution, $p(a), p(b)$ are marginal distributions, and $I(;)$ denotes MI. Higher MI values indicate stronger relationships.

\begin{equation}
    I(A;B) = \sum_{a \in \mathcal{A}} \sum_{b \in \mathcal{B}} p(a,b)  \left(\frac{p(a,b)}{p(a)p(b)}\right)
    \label{eq:MI}
\end{equation}

\textbf{MI-based extracted information evaluation} 

Given a response sequence $R$ with total token length $n$, let $Y_{l} \in \mathbb{R}^{n \times d}$ denote the MHA embedding at layer $l$ with hidden dimension $d$. The extraction strategy $\pi$ selects $k$ tokens for each layer $l$ and returns the output embedding as $O_l$.

\begin{equation}
O_l = \pi(Y_l) \in \mathbb{R}^{k \times d}
\end{equation}

The whole process of contribution analysis can be simplified as an Extract-then-classify problem. Inspired by Information Bottleneck, we try to find the optimal trade-off between compression (minimizing redundant information between $O_l$ and $Y_l$), and prediction (preserving relevant information between $O_l$ and classification result $C$). Additionally, $k$ tokens are selected to form $O_l$, and redundancy between themselves should also be minimized. Therefore, the final metric $K_{IB}$ is defined in Eq. \ref{eq:Mutual_Information}, \ref{eq:MI_eq2} and \ref{eq:MI_eq3}, where $\beta$ is set to 0.5.

\begin{equation}
\begin{gathered}
\small
K_{IB} = \sum_{l} ( I(C; O_l) - (1-\beta) I(Y_l; O_l) - \\
         \beta \sum\limits_{\substack{1 \leq n_1 \leq n_2 \leq k}} I(O_{n_1,l}; O_{n_2,l}) )
\label{eq:Mutual_Information}
\end{gathered}
\end{equation}

\begin{equation}
I(C;O_l) = \sum_{k_i=1}^{k}I(C;O_{k_i,l})
\label{eq:MI_eq2}
\end{equation}

\begin{equation}
I(Y_l;O_l) =  \sum_{k_i=1}^{k}I(Y_l;O_{k_i,l})
\label{eq:MI_eq3}
\end{equation}

$K_{IB}$ estimates the valid information maintained throughout the process. Higher $K_{IB}$ means better extraction strategy $\pi$.

\subsection{Contrastive-learning based Non-copyright Filtering Module}
We employ unsupervised learning exclusively on copyrighted datasets to train the non-copyright response filtering module, because non-copyrighted texts are diverse and challenging to collect comprehensively. Therefore, a contrastive-learning-based non-copyright filtering module is proposed. 

\textbf{Utilizing Sub-dataset Prior in LLM Internal Representations}

Our framework builds upon two key observations: (1) copyrighted datasets inherently contain rich structural information in their sub-categories (2) Such information lies in LLM internal representations, serving as a prior to enhance non-copyright response filtering. 

\begin{figure}[htbp]
    \centering
    \includegraphics[width=\linewidth]{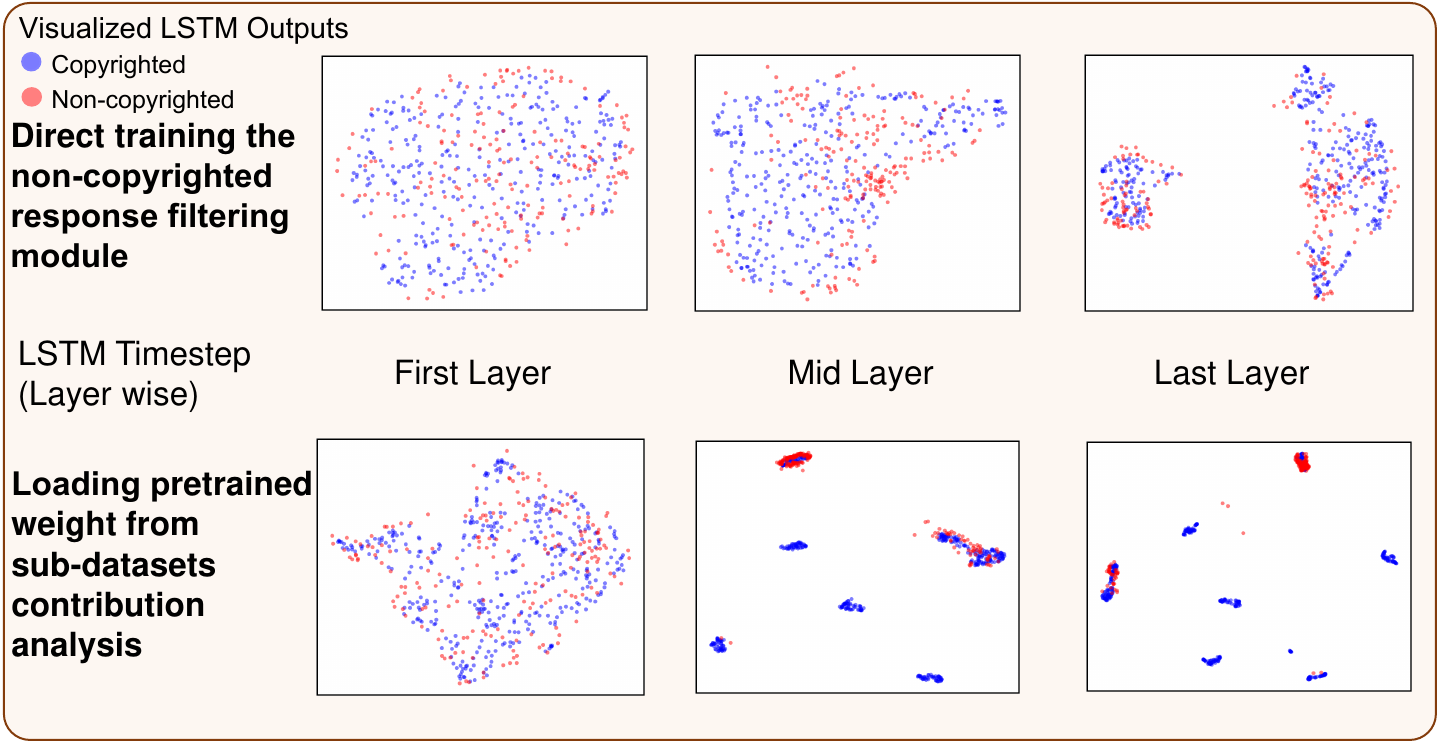} 
    \caption{Visual comparison (UMAP) of the effect of applying contribution analysis prior. Top Row: LSTM module outputs in different timesteps through direct training of a filtering module. Bottom Row: LSTM module outputs in different timesteps with pre-trained prior. }
    \label{fig:prior}
\end{figure}

A preliminary visualization is shown in Fig. \ref{fig:prior}. Copyrighted data (blue dots) and non-copyrighted ones (red dots) are fed into the LSTM-based feature extractor module, and the layer-wise output in each time step is visualized through UMAP. Compared with the top row, the bottom row in Fig. \ref{fig:prior} shows that pre-trained LSTM gradually separates intra-class copyrighted datasets even without training, efficiently reducing overlap between copyright and non-copyright data. Therefore, we conclude such insights with the following procedures:

\begin{itemize}
    \item \textit{Structure-aware Local Prior Pretraining.} Based on the trained copyrighted sub-datasets contribution analysis module, given token representation $O \in \mathbb{R}^{L \times k \times d}$, an LSTM-based feature extractor f: $\mathbb{R}^{k \times d} \rightarrow \mathbb{R}^c$ is equipped the capability to learn localized embeddings.
    \item \textit{Global Space Fine-tuning.} A global projector is further fine-tuned to learn a high-dimensional space
    for further copyright/non-copyright decisions.
\end{itemize}

\textbf{Network Design based on Contrastive Learning}

We achieve non-copyright response filtering by training a module that directly detects non-copyright contexts from $D$. Copyright datasets $D_{c}$ are batched into $\{T_{b}\}_{b=1}^{\mid B \mid}$, and augmented into $\{\tilde{T_{b}}\}_{b=1}^{\mid B \mid}$ using Random Span Masking (RSM) for data augmentation \cite{liu2021fast, cho2022enhancing}. Thus, the final embeddings of such pairs are denoted as $e_b$ and $\tilde e_b$. These augmented pairs are passed through a frozen LLM to generate layer-wise MHA outputs, which are processed by the LSTM-based feature extractor and the copyrighted sub-datasets contribution analysis module. In this module, the global projector (lightweight MLP) processes the first-layer MLP outputs from the data pairs, and produces two embeddings, $e_b$, $\tilde e_b$. The module is trained using contrastive loss, which encourages similar embeddings of positive pairs while separating embeddings of different samples. Non-copyright issues can then be detected by thresholding in the learned embedding space, as shown in Fig. \ref{fig:whol}(c). The loss for each query text index $i$ is computed as in Eq. \ref{eq:loss}, where $\tau$ is the temperature parameter.

\begin{equation}
L_{\mathrm{CL}} = -\sum_{i=1}^{\mid B \mid} \log \left( \sum_{j = 1 }^{\mid B \mid} \frac{\exp \left(\boldsymbol{e}_i \cdot \tilde{\boldsymbol{e}}_j / \tau\right)}{\sum_{k=1, k \neq i}^{\mid B \mid} \exp \left(\boldsymbol{e}_i \cdot \tilde{\boldsymbol{e}}_k/ \tau\right)} \right)
\label{eq:loss}
\end{equation}

\textbf{Non-copyright Filtering Policy}

We leverage the embedding space learned from copyright data to filter out non-copyright responses based on distance-based measurement. Specifically, we use Mahalanobis distance, a classic distance-based scoring function $d_M$ \cite{lee2018simple}, which calculates the distance between a vector point $\mathbf{x}$ and a distribution $\mathbf{R_Q} \in \mathbb{R}^N$, as defined in Eq. \ref{eq:M_distance}. Here, $\mathbf{x}$ represents embedding of LLM response sample, while $\boldsymbol{\mu}$ and $\mathbf{\Sigma}$ represents mean and positive semi-definite covariance matrix of embeddings from distribution of copyrighted responses $\mathbf{R_Q}$, and $M$ is the dimensionality of $\boldsymbol{\mu}$. As in Eq. \ref{eq:decision}, the anomaly threshold $\delta$ determines whether a text is non-copyright. This $\delta$ is set to ensure a 95\% true positive rate (TPR) on the training samples.

\begin{equation}
{d}_{{M}}(\mathbf{x}, \mathbf{R_Q})=\sum_{i=1}^M\left(\mathbf{x_i}-\boldsymbol{\mu}_{}\right) \mathbf{\Sigma}_{}^{-1}\left(\mathbf{x_i}-\boldsymbol{\mu}_{}\right)^T
\label{eq:M_distance}
\end{equation}

\subsection{Controllable Text Generation Design}
\label{sec:prompt}

A remaining challenge in evaluating the performance of \sysname{} is collecting the desired LLM responses that suffer copyright issues.
Moreover, the ground truth of which specific copyright is infringed in response is also unknown. To address the problem, we propose controllable text generation designs based on copyright-related prompts. Our method first draws inspiration from established copyright infringement criteria for books \cite{wikipediaSubstantialSimilarity}, and extends to analyze broader datasets in the Pile via three generation scenes and text mixture task.

\textbf{Generation of Real-World Copyright-infringed Texts in Books}

Copyright law uses the `substantial similarity' criteria to identify infringement in books. This criterion examines similarities in unique expressions, plot elements, character relationships, and settings between works \cite{wikipediaSubstantialSimilarity}.

\begin{figure}[htbp]
    \centering
    \includegraphics[width=\linewidth]{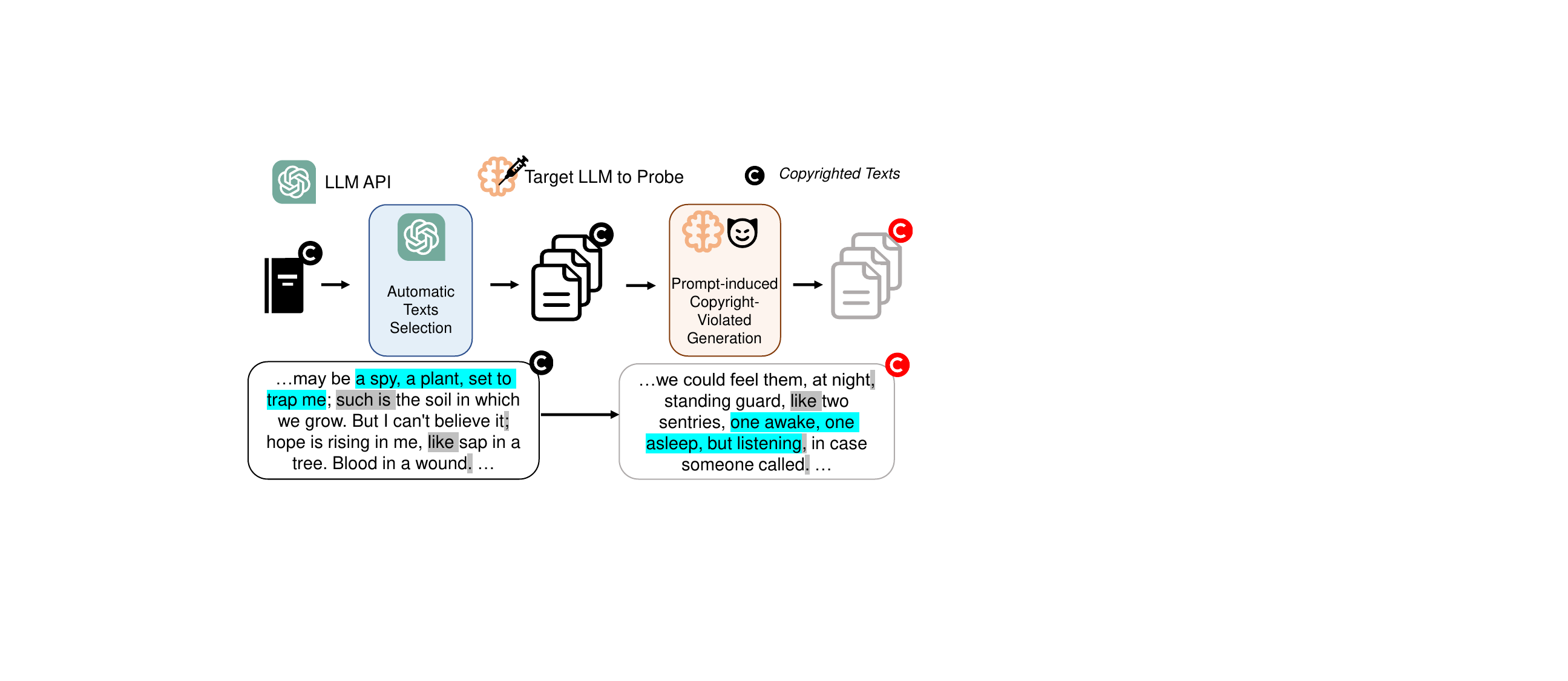} 
    \caption{Workflow of generating real-world copyright-infringed texts. An example shows a certain amount of substantial similarity exists between the original book and generated texts.}
    \label{fig:copyrightbook}
\end{figure}

Based on this, we design a controlled generation process for books using LLMs' language understanding ability. By incorporating these criteria into prompt templates and providing copyrighted source text fragments, we guide LLMs to generate texts with deterministic copyright-infringement to the original works. To extract relevant text fragments from copyrighted authors, we use the Kimi API from Moonshot AI  \cite{Qin2024MooncakeAK}, with a complete process shown in Fig. \ref{fig:copyrightbook}.

\textbf{Text Generation Scenes and Mixture Task}

\begin{figure}[htbp]
    \centering
    \includegraphics[width=\linewidth]{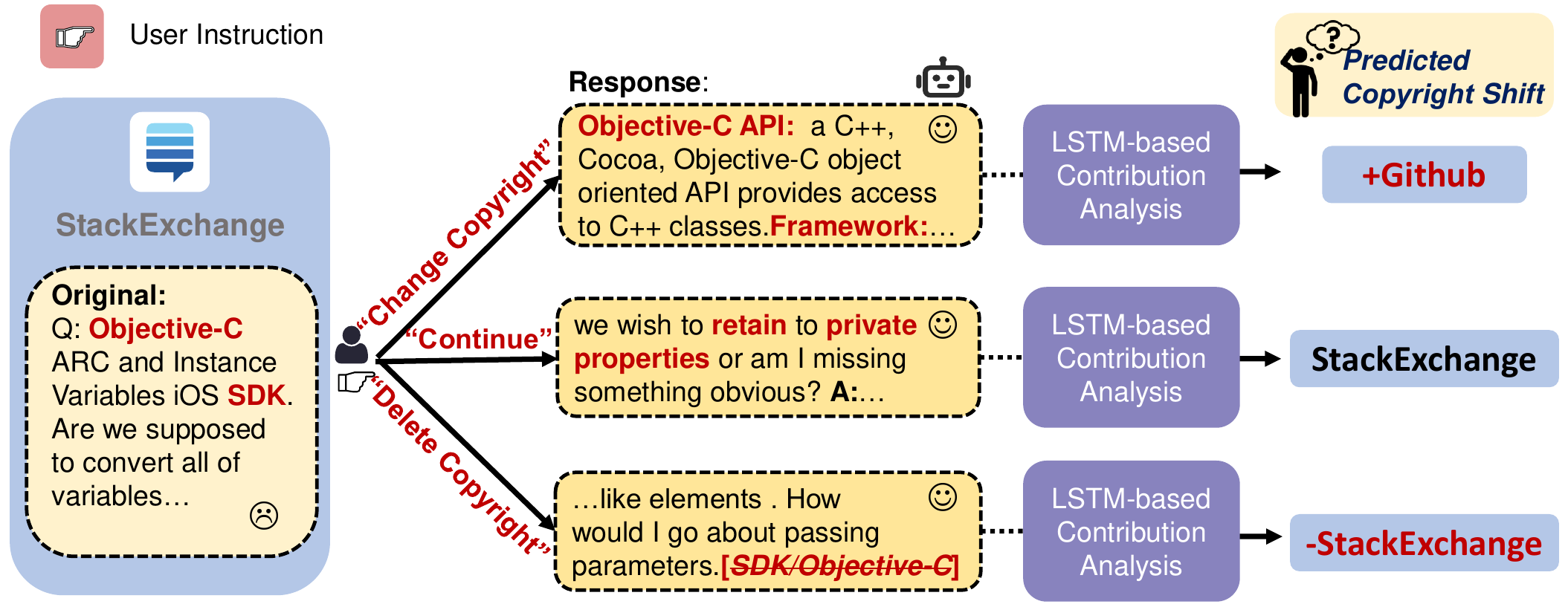} 
    \caption{Overview of three controlled generation scenes: text continuation, copyright transfer, and copyright elimination. Shifts in dataset contributions due to user editing instructions are expected to be detected. } 
    \label{fig:scenes_main_prompt}
\end{figure}

Criteria for books cannot be directly applied to other datasets. Therefore, to extend to broader datasets in the Pile \cite{gao2020pile}, we design three controlled text generation scenes: continuation, copyright-transfer, and copyright-elimination (Fig. \ref{fig:scenes_main_prompt}). Each scene represents a distinct way of generating potentially copyright-related content.

For continuation, the text is directly generated from the input. For copyright transfer, we design prompt templates to ask LLM to rewrite texts in the style of specific target datasets. The copyright-elimination scene aims to ask LLM to remove distinctive stylistic features from the texts.

To further evaluate the contribution analysis module in \sysname{} quantitatively, a text mixture task is designed, where input texts into the framework are a mixture of multiple copyrighted training datasets with predefined portions.

\section{Case Study: Attributing Real-world Copyright-infringed LLM Generated Texts }

The Books3 subset in the Pile \cite{gao2020pile} has been proven to contain copyrighted books from famous authors and used for LLM training, such as GPT-Neo and GPT-J \cite{Neo, gpt-j}. Our task is to determine which author contributes the most if the generated text has a copyright violation. In this section, we show \sysname{} is more effective in such analysis in real-world scenarios than other methods.

\begin{table}[ht]
    \centering
    \scriptsize
    \caption{Copyrighted books identified in Books3 dataset for copyright analysis in case study.}
    \begin{tabular}{@{}l@{\hspace{10mm}}p{4.5cm}@{\hspace{10mm}}l@{}}
        \toprule
        Author       & Book Title   & Year    \\ 
        \midrule
        Stephen King     & Pet Sematary   & 1983 \\
        Haruki Murakami     &  1Q84 &  2009 \\
        James Patterson     & The Midnight Club & 1999   \\
        John MacArthur     & Twelve Unlikely Heroes  &  2012  \\
        Donald J. Trump      & Great Again: How to Fix Our Crippled America & 2015  \\
        Eugenio Amato     & Law and Ethics in Greek and Roman Declamation & 2015   \\
        Jules Brown     & The Rough Guide to the Lake District & 2000   \\
        Rita Leganski     & The Silence of Bonaventure Arrow & 2012   \\
        \bottomrule
    \end{tabular}

    \label{tab:authors_books}
\end{table}

\subsection{Experiment Setup}

\textbf{Data Preparation}

We choose eight books from representative copyrighted authors, \cite{theguardianZadieSmith}, which are included in Books3 \cite{gao2020pile}. The full list is shown in Tab. \ref{tab:authors_books}. 
We focus on developing a method that can effectively work with limited training data, which is crucial for real-world copyright protection scenarios where extensive training samples from each author may not be readily available or practical to obtain. Thus, we randomly extract 160 snippets of 512 tokens from author's works for training, and 4800 snippets for validation. This formulates an eight-class classification problem where the model learns to distinguish between authors based on their original copyrighted texts.

To evaluate \sysname{}'s ability to classify in real-world copyright-infringed texts, we construct a test set by selecting representative text snippets, and then prompt target LLM for text generation. We first select 400 text snippets that embody their distinctive themes or writing styles using the Kimi API \cite{Qin2024MooncakeAK}. These snippets are then processed in target LLM using our prompt engineering method (sec. \ref{sec:prompt}) to generate infringed texts. The model's classification accuracy on these generated samples directly measures its capability to trace LLM-generated texts back to the original authors. Specifically, higher accuracy indicates better performance in identifying which author's work has been infringed by the LLM.

We also conduct non-copyright response filtering task, where 900 non-copyrighted texts are collected from Gutenberg \cite{gutenbergProjectGutenberg}, and 900 copyrighted texts from Books3.

\textbf{Baseline Settings}

We propose two baselines for the copyright sub-datasets contribution analysis. The prompt engineering baseline ($Baseline_1$/B1) is designed based on templates in \cite{person}, treating LLMs as agents to ask them to decide the sub-dataset contributions according to the given texts. For $Baseline_2$ (B2), we implement a distance-based approach to show the importance of layer-wise dependencies. This method computes class-specific representative embeddings by averaging MHA across all layers of training samples within each class. For classification, the test sample is assigned to the class whose representative has the shortest L1 distance to the sample's embedding. Thus, $Baseline2$ uses LLM architecture but ignores layer-wise patterns.

Besides prompt-based attribution and distance-based methods, text classification methods (BERT \cite{kenton2019bert} and RoBERTa \cite{liu2019roberta} both with full parameter finetuning) are compared simultaneously.

\subsection{Results and Analysis}
We evaluate all methods on both copyrighted books and LLM-generated texts. For supervised methods, models are trained on copyrighted books. Meanwhile, prompt-based attribution directly queries target LLM by prompt templates. On the evaluation of copyrighted books, standard text classification methods and our proposed methods both achieve over 95\% accuracy. The remaining methods (prompt attribution and distance-based methods) perform significantly worse, all below 40\%, as shown in Tab. \ref{tab:book-acc}. 

To further distinguish text classification methods and 
\sysname{}, we notice when evaluating on LLM-generated copyrighted texts, \sysname{} (shown in bold in the 3nd column in Tab. \ref{tab:book-acc}) consistently outperforms other methods by over 8.0\% in accuracy. This is because \sysname{} effectively utilizes MHA features, which better correlates the distributional shifts between original copyrighted books and LLM-generate texts. We also find our method performs well on non-copyright response filtering in Tab. \ref{tab:case_study_ood}.

\begin{table}[h!]
    \centering
    \footnotesize
    \caption{Classification accuracy of different methods in the case study. GPT-Neo-2.7B generates copyright-infringed texts, and serves as target LLM for \sysname{}, prompt attribution, and distance-based methods.}
    \setlength{\tabcolsep}{3.5pt}
    \begin{tabular}{@{}l@{\hspace{5pt}}c@{\hspace{5pt}}c@{\hspace{5pt}}c@{}}
        \toprule
        \textbf{Method} & 
        \makecell[c]{\textbf{Accuracy(\%)$\uparrow$}\\ Original Book} & 
        \makecell[c]{\textbf{Accuracy(\%)$\uparrow$}\\ \textbf{Generated Text}} &
        \makecell[c]{\textbf{Cost(Min/GB)$\downarrow$}\\ Time/Memory} \\
        \midrule
        Inner-Probe (Inter) & \textbf{98.5} & \textbf{72.7} & \textbf{8.0/5.9}\\
        Inner-Probe (Var) & 97.1 & \textbf{78.7}  & \textbf{5.9/7.2}\\
        Inner-Probe (A-Var) & 95.5 & \textbf{80.7} & \textbf{5.5/7.2}\\
        BERT \cite{kenton2019bert} & 98.1 & 72.7 & 29.4/58.3 \\
        RoBERTa \cite{liu2019roberta} & 96.1 & 72.5 & 29.5/58.8\\
        Prompt Attribution (B1) & 9.8 & 8.9 & 27.6/46.8\\
        Distance-based (B2) & 32.5 & 24.5 & 0.5/CPU\\
        \bottomrule
    \end{tabular}
    \label{tab:book-acc}
\end{table}

\begin{table}[h]
    \centering
    \footnotesize
    \setlength{\tabcolsep}{3.5pt}
    \caption{Non-copyright Dataset Detection Accuracy and AUC Comparison on GPT-Neo-2.7B in the case study.} 
    \begin{tabular}{@{}lccccccc@{}}
        \toprule
        Method & Inter.  & Var. &  A-Var. &  Baseline \cite{cho2022enhancing} \\ 
        \midrule
        Accuracy(\%)($\uparrow$) & 83.75 & 94.95 & 86.75 & 79.60 \\

        AUC($\uparrow$) & 0.951 & \textbf{0.982} & 0.966 & 0.943 \\
        \bottomrule
    \end{tabular}
    \label{tab:case_study_ood}
\end{table}

\section{Experiment}

\subsection{Overview}

Following the Books3 dataset in the case study, we first evaluate \sysname{} on its parent corpus, the Pile \cite{gao2020pile}, in the main experiments. These datasets are widely used and known for copyright issues in LlaMA and GPT pretraining \cite{touvron2023llama, detecting, das2025blind, tao2025evidencing}, yet they offer limited coverage and only partial transparency for disclosed LLMs. To expand coverage, we add analysis on the larger, multilingual dataset called Matrix. To improve transparency, we use an open-source LLM with a public pretraining pipeline and dataset, namely Map-Neo 7B trained on Matrix \cite{zhang2024map}.

Therefore, we report results in two tracks. Besides the main experiments, Section \ref{sec:extend} first extends experiments to the Matrix dataset, then discusses limitations by studying how LLM generation length and data scale affect \sysname{}.

\subsection{Experiment Setup}

\textbf{Model and Datasets}

We test our methods over three open-source language models, including encoder-based and decoder-based ones: bert-base-uncased \cite{bert-base-uncased}, GPT-Neo-2.7B \cite{Neo} and GPT-J-6B \cite{gpt-j}. 

For the encoder-only model, we evaluate six GLUE classification sub-datasets: CoLA, SST-2, MRPC, WNLI, QQP, and RTE, with each sub-dataset comprising 500 training and 100 testing samples. Before further experiments, BERT is fine-tuned on tasks in GLUE. For the decoder-based models, experiments utilize the Pile dataset, an 825 GB open-source collection of 22 diverse sub-datasets \cite{gao2020pile}. 8 classes including Github, OpenWebText2, Wikipedia (en), StackExchange, PubMed Abstracts, Pile-CC, USPTO Backgrounds, and FreeLaw are considered copyrighted datasets, and 4 classes including PubMed Central, Enron Emails, OpenSubtitles, and DM Mathematics are considered non-copyrighted ones.

\textbf{Baseline Settings}

For the copyright sub-datasets contribution analysis module, we compare our method with text classification models, including BERT \cite{kenton2019bert}, and RoBERTa \cite{liu2019roberta}.
We also implement a prompt-based method ($Baseline1$/B1) and a distance-based method ($Baseline2$/B2).

For the non-copyrighted responses filtering module, we compare with the unsupervised method \cite{cho2022enhancing} as the baseline. We evaluate it under two settings for fair comparisons: (\textit{Setting1}) with copyright sub-category information incorporated through an extra cross-entropy loss implemented in \cite{cho2022enhancing}. (\textit{Setting2}) treating all copyright data as a single class in unsupervised learning. Our method uses \textit{Setting2}, and both settings are used in the baseline.

\subsection{Copyright Sub-datasets Contribution Analysis Module}

\subsubsection{Classification Performance on Original Copyrighted Texts}

\textit{Data Preparation}
We allocate up to 12,800 train and 6,400 test samples per copyrighted class. Evaluations are done on the maximum test set.

\textbf{Classification Performance Comparison}

We compare classification performance on three language models: BERT, GPT-Neo-2.7B, and GPT-J-6B. Results of text classification methods remain the same for each model, because they do not utilize LLM itself. 

\begin{figure*}[htbp]
    \centering
    \includegraphics[width=\linewidth]{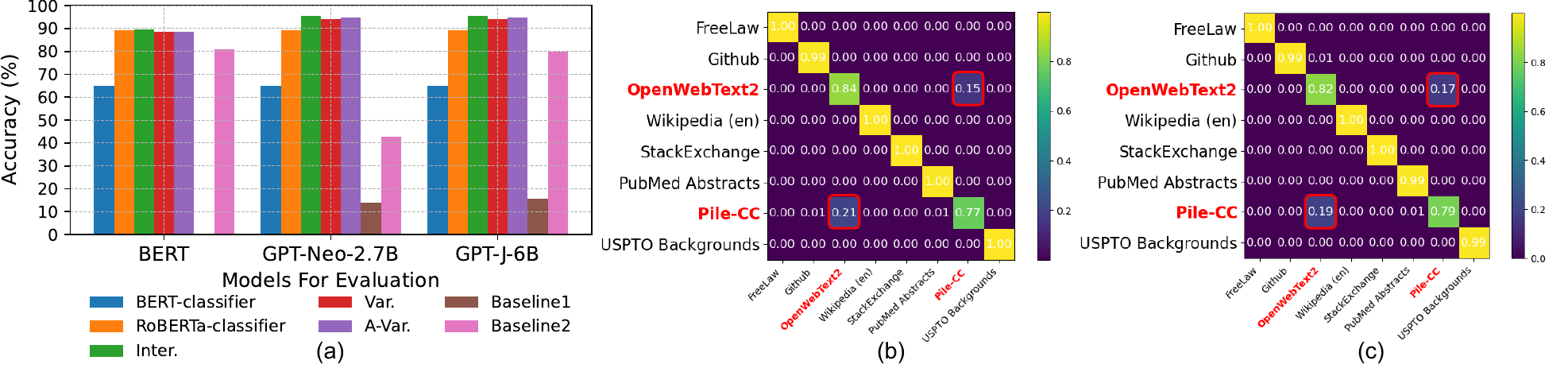} 
    \caption{Classification performance comparison across various methods and language models. (a) Original copyright data classification performance comparison. Self-correlation analysis for sub-dataset inference on (b) GPT-Neo-2.7B and (c) GPT-J-6B.} 
    \label{fig:overall}
\end{figure*}

As shown in Fig. \ref{fig:overall}(a), \sysname{} achieves contribution analysis accuracies of 95.4\%, 94.3\%, and 94.9\% respectively on GPT-Neo-2.7B (\textsc{Interval}, \textsc{Var}, and \textsc{A-Var}), outperforming text classification methods (best 89.27\%) and other two baselines (best 79.9\%) by 6.13\% accuracy. Despite $Baseline_2$ achieving over 80\% accuracy in BERT, it drops to 40\% in GPT-Neo-2.7B and stays below 80\% in GPT-J-6B. In contrast, \sysname{} shows steady accuracy above 90\%  in each model. Similar results are in BERT and GPT-J-6B.

\begin{figure*}[htbp]
    \centering
    \includegraphics[width=\linewidth]{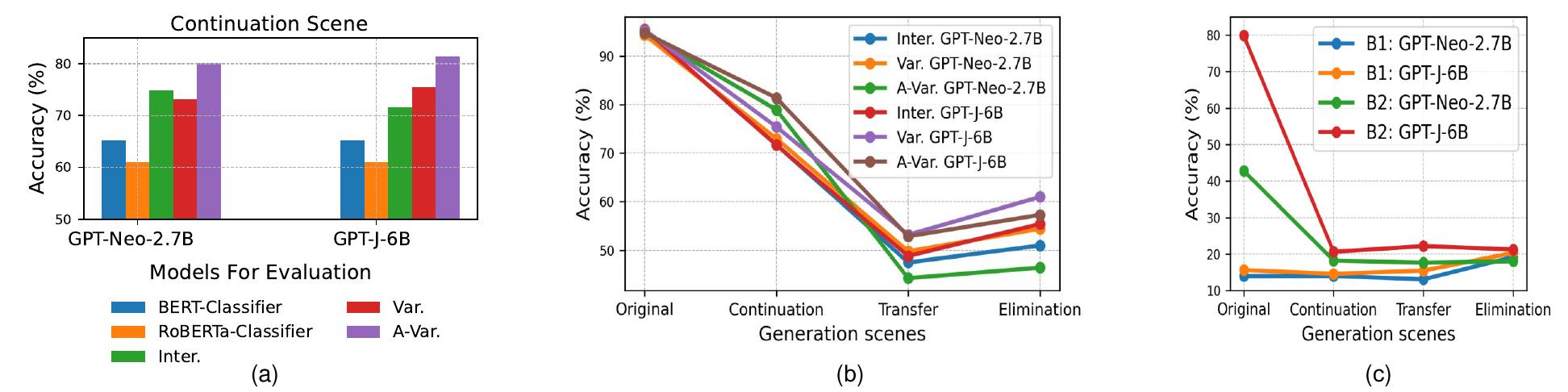} 
    \caption{Comparisons of copyright contribution analysis methods in different generation scenes. (a) Accuracy comparisons with text classifiers in continuation scene (b) Overall performance of our methods in generation scenes (c) Overall performance of Baseline1 and Baseline2 in generation scenes.}
    \label{fig:scenes}
\end{figure*}

\textbf{Self-correlation Predictions in Datasets}

Correlations naturally occur between sub-datasets, and our method effectively captures this. Using our method, we calculate the average probability distributions for each class during inference. As shown in Fig. \ref{fig:overall}(b) and (c), GPT-Neo-2.7B and GPT-J-6B reveal similar correlations between OpenWebText2 and Pile-CC, which are expected as both are web-sourced components. The consistent result across models validates the ability of our method.

\subsubsection{Evaluation on Controlled Generation Scenes}

\textit{Data Preparation}
For the continued generation scene, we let LLM continue to generate based on 6400 test samples used before. In copyright transfer, all related elements are transferred with "Github" stylistic elements and removed during copyright elimination.

\textbf{Evaluation on Continual Generation Scene}

We evaluate our method with text classification baselines on the continual generation scene, where LLM-generated texts naturally inherit the original copyright category. Therefore, higher classification accuracy indicates better contribution analysis on LLM-generated texts. As shown in Fig. \ref{fig:scenes}(a), our methods show accuracies of 74.86\%, 73.2\%, and 80.19\% separately on GPT-Neo-2.7B, still outperforming text classification (with the best accuracy 65.15\%) by 15.04\%.

Although \textsc{Inter} outperforms variance-based methods in Fig. \ref{fig:overall}, variance-based methods significantly improve accuracy in larger models, whereas \textsc{Inter} shows minimal improvements, as shown in Fig. \ref{fig:scenes}(a). What's more, \textsc{A-Var} outperforms \textsc{Inter} and \textsc{Var}, which demonstrates robustness and generalization.

\textbf{Other Scenes and Text Mixture Task}

We measure other generation scenes from a qualitative perspective, which is shown in Fig. \ref{fig:scenes}(b)(c). We notice all methods show an accuracy shift when texts change from continuation to transfer, and when in the copyright elimination scene, most methods tend to make random guesses. 

\begin{figure*}[htbp]
    \centering
    \includegraphics[width=\linewidth]{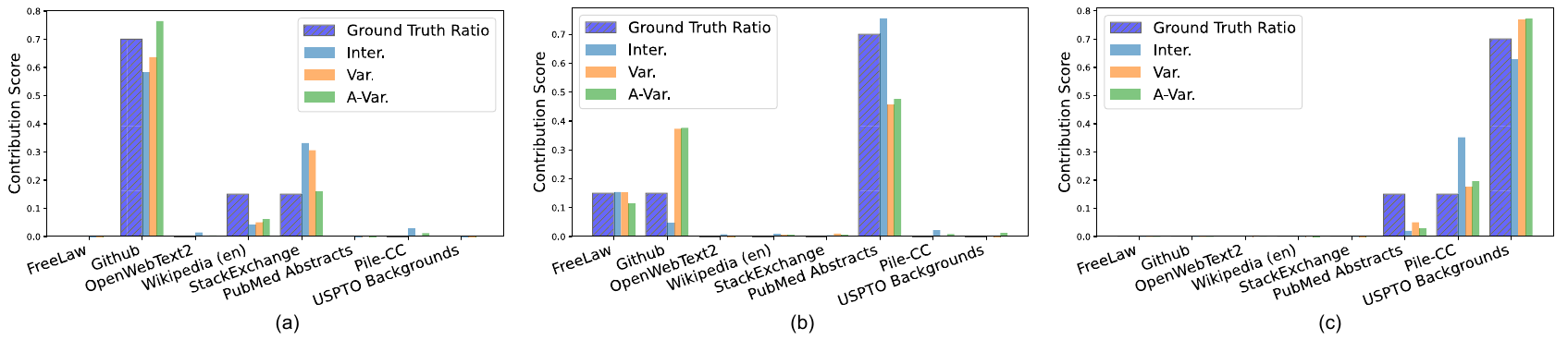} 
    \caption{Comparison of predicted copyright contribution scores with ground truth ratios (15\%, 15\%, 70\%) in text mixture task. Three random mixture examples, (a), (b), and (c), are shown to illustrate effectiveness.} 
    \label{fig:mix}
\end{figure*}

For text mixture task, all methods in \sysname{} can effectively detect mixture ratios with minimal loss. We randomly select 3 classes from 8 copyrighted datasets, and mix them into new samples by taking 15\%, 15\%, and 70\% of the token lengths respectively. As three examples shown in Fig. \ref{fig:mix}, predicted contribution from \textsc{Inter}, \textsc{Var} and \textsc{A-Var} proposed in \sysname{} aligns closely with ground truth ratios. We further compute MSE loss between predicted contribution scores and predefined portions, which is lower than 0.05 on average.

\textbf{Explanation by Mutual Information}

We conduct experiments on GPT-Neo-2.7B and GPT-J-6B, and find that $K_{IB}$ accurately predicts the classification accuracy. As depicted in Tab. \ref{Tab:IB_2.7B}, $K_{IB}$  correlates with classification accuracy with exactly the same trend. The interval-based method shows better IB metrics, which corresponds to its higher accuracy. 

In conclusion, we design a metric that quantitatively explains each extraction method and paves the way for optimized designs.

\begin{table}[h!]
\captionsetup{aboveskip=10pt, belowskip=10pt}
\centering
\footnotesize
\caption{Explaining the superior performance of  \textsc{Inter} in \sysname{} using Mutual Information analysis on GPT-Neo-2.7B. Notably, the $K_{IB}$ results demonstrate a consistent trend with classification accuracy. Results of GPT-J-6B are not shown for space limit.\label{Tab:IB_2.7B}}
\begin{tabular}{lccc}
\toprule
Methods & \textsc{Inter.} & \textsc{Var.} & \textsc{A-Var.} \\
\midrule
  Classification Accuracy (\%) & \cellcolor[gray]{0.6}85.5 & \cellcolor[gray]{1.0}76.9 & \cellcolor[gray]{0.8}83.8 \\
\midrule
\textbf{Mutual Information (MI)} & & & \\
$I(Y;O_1)(\downarrow)$ & \cellcolor[gray]{0.6}-6.86 & \cellcolor[gray]{1.0}11.09 & \cellcolor[gray]{0.8}10.13 \\
$I(Y;O_2)(\downarrow)$ & \cellcolor[gray]{0.6}-1.63 & \cellcolor[gray]{1.0}12.63 & \cellcolor[gray]{0.8}12.46 \\
$I(Y;O_3)(\downarrow)$ & \cellcolor[gray]{0.6}1.96 & \cellcolor[gray]{0.8}12.87 & \cellcolor[gray]{1.0}12.88 \\
$I(O_1;O_2)(\downarrow)$ & \cellcolor[gray]{0.6}1.52 & \cellcolor[gray]{1.0}42.87 & \cellcolor[gray]{0.8}28.79 \\
$I(O_1;O_3)(\downarrow)$ & \cellcolor[gray]{0.6}1.50 & \cellcolor[gray]{1.0}38.55 & \cellcolor[gray]{0.8}23.26 \\
$I(O_2;O_3)(\downarrow)$ & \cellcolor[gray]{0.6}18.04 & \cellcolor[gray]{1.0}42.68 & \cellcolor[gray]{0.8}27.59 \\
$I(C;O_1)(\uparrow)$ & \cellcolor[gray]{0.6}36.44 & \cellcolor[gray]{1.0}1.17 & \cellcolor[gray]{0.8}2.78 \\
$I(C;O_2)(\uparrow)$ & \cellcolor[gray]{1.0}0.17 & \cellcolor[gray]{0.8}0.39 & \cellcolor[gray]{0.6}0.63 \\
$I(C;O_3)(\uparrow)$ & \cellcolor[gray]{0.6}1.44 & \cellcolor[gray]{1.0}0.31 & \cellcolor[gray]{0.8}0.65 \\
$I(Y;O)(\downarrow)$ & \cellcolor[gray]{0.6}-6.53 & \cellcolor[gray]{1.0}36.59 & \cellcolor[gray]{0.8}35.46 \\
\midrule
 \textbf{$K_{IB}$($\uparrow$)} & \cellcolor[gray]{0.6}30.79 & \cellcolor[gray]{1.0}-78.47 & \cellcolor[gray]{0.8}-53.50 \\
\bottomrule
\end{tabular}

\end{table}

\subsection{Non-copyrighted Responses Filtering Module}

\textit{Data preparation} We construct the training dataset with 800 copyrighted samples for contrastive learning, and a test set with 400 copyrighted and 200 non-copyrighted samples. The copyrighted samples have 8 classes. To ensure fairness, both methods use Random Span Masking (RSM) for data augmentation. The filtering threshold is set to maintain a 95\% true positive rate (TPR).

Incorporating intra-class copyright information improves performance, but using LSTM prior learnt from LLM has more to offer compared with using it during the unsupervised learning stage. As in Tab. \ref{tab:OOD}, baseline \cite{cho2022enhancing} with \textit{Setting1} improves accuracy by 3.94\% and AUC by 0.108, yet still underperforms our method (\textit{Setting2}) by an AUC of 0.104. Our method (\textsc{Inter}, \textsc{Var} and \textsc{A-Var}) shows consistent advantages and is more computationally efficient, as it doesn't require fine-tuning on language models like BERT.

\begin{table}[h]
    \centering
    \footnotesize
    \setlength{\tabcolsep}{2.6pt}
    \caption{Non-copyright Detection Accuracy and AUC Results.} 
    \begin{tabular}{@{}l*{8}{c}@{}}
        \toprule
        & \multicolumn{2}{c}{Inter.} & \multicolumn{2}{c}{Var.} & \multicolumn{2}{c}{A-Var.} & \multicolumn{2}{c}{Baseline \cite{cho2022enhancing}} \\ 
        \cmidrule(lr){2-3} \cmidrule(lr){4-5} \cmidrule(lr){6-7} \cmidrule(lr){8-9}
        Model Size & 2.7B & 6B & 2.7B & 6B & 2.7B & 6B & \textit{Setting1} & \textit{Setting2} \\
        \midrule
        Accuracy(\%)($\uparrow$) & 92.97 & 90.80 & 88.95 & 93.83 & 83.92 & 88.33 & 70.31 & 66.37 \\
        AUC($\uparrow$) & \textbf{0.954} & 0.934 & 0.917 & 0.953 & 0.886 & 0.935 & 0.850 & 0.742 \\
        \bottomrule
    \end{tabular}
    \label{tab:OOD}
\end{table}

\subsection{Training Efficiency and Data Scale Effect}

\textit{Robustness to Data Scale} Large amounts of copyrighted data pose challenges to model training, necessitating a lightweight method. To evaluate data scaling effects, we test \sysname{} by setting maximum token number $k$ to 7, and varying training samples from 160 to 1,600 and 12,800, with 6,400 inference samples. As shown in Tab. \ref{tab:datascale}, training on 1,600 samples is sufficient for \sysname{} to outperform baselines, while reducing training time (Tab. \ref{tab:model_efficiency}). The accuracy increases linearly with data scale and $k$. Our empirical study shows that while a small $k$ is generally sufficient, $k=7$ leads to model degradation.

\begin{table}[htbp]
\centering
\footnotesize
\setlength{\tabcolsep}{2pt} 
\caption{Data scale effect across different methods and models. $k$ denotes the sampling token number. The model accuracy strictly follows the increase in $k$ and data scale until $k = 7$. }
\label{tab:datascale}
\begin{tabular}{@{}lccccccccccc@{}}
\toprule
Data scale & $k$ & \multicolumn{2}{c}{Inter.} & \multicolumn{2}{c}{Var.} & \multicolumn{2}{c}{A-Var.} & \multicolumn{2}{c}{Baseline1} & \multicolumn{2}{c}{Baseline2} \\
\cmidrule(lr){3-4} \cmidrule(lr){5-6} \cmidrule(lr){7-8} \cmidrule(lr){9-10} \cmidrule(lr){11-12} 
& & 2.7B & 6B & 2.7B & 6B & 2.7B & 6B & 2.7B & 6B & 2.7B & 6B\\
\midrule
\multirow{4}{*}{\makecell[l]{Small\\(160,6400)}} 
& 1 & 69.6 & 75.6 & 63.4 & 75.7 & 75.4 & 70.8 & \multirow{4}{*}{14.0} & \multirow{4}{*}{15.6} & \multirow{4}{*}{39.2} & \multirow{4}{*}{72.3} \\
& 3 & 85.5 & 87.1 & 76.9 & 83.5 & 83.8 & 80.4 & & & & \\
& \textbf{5} & \textbf{85.9} & \textbf{88.6} & \textbf{79.7} & \textbf{85.9} & \textbf{85.9} & \textbf{82.7} & & & & \\
& \textcolor{gray}{7} & \textcolor{gray}{82.7} & \textcolor{gray}{87.6} & \textcolor{gray}{67.4} & \textcolor{gray}{84.7} & \textcolor{gray}{79.0} & \textcolor{gray}{81.4} & & & & \\
\midrule
\multirow{4}{*}{\makecell[l]{Medium\\(1600,6400)}} 
& 1 & 88.6 & 89.0 & 89.9 & 91.2 & 89.5 & 88.3 & \multirow{4}{*}{14.0} & \multirow{4}{*}{15.6} & \multirow{4}{*}{42.7} & \multirow{4}{*}{79.4} \\
& 3 & 92.8 & 93.1 & 91.5 & 92.6 & 91.8 & 91.1 & & & & \\
& \textbf{5} & \textbf{93.3} & \textbf{93.6} & \textbf{92.3} & \textbf{93.2} & \textbf{93.1} & \textbf{91.3} & & & & \\
& \textcolor{gray}{7} & \textcolor{gray}{93.4} & \textcolor{gray}{94.0} & \textcolor{gray}{92.5} & \textcolor{gray}{93.2} & \textcolor{gray}{93.0} & \textcolor{gray}{91.7} & & & & \\
\midrule
\multirow{4}{*}{\makecell[l]{Large\\(12800,6400)}} 
& 1 & 92.0 & 93.3 & 93.4 & 93.9 & 92.2 & 93.3 & \multirow{4}{*}{14.0} & \multirow{4}{*}{15.6} & \multirow{4}{*}{42.8} & \multirow{4}{*}{79.9} \\
& 3 & 94.8 & 95.4 & 94.3 & 95.3 & 94.3 & 94.4 & & & & \\
& \textbf{5} & \textbf{95.4} & \textbf{95.5} & \textbf{94.3} & \textbf{95.4} & \textbf{94.9} & \textbf{94.7} & & & & \\
& \textcolor{gray}{7} & \textcolor{gray}{95.2} & \textcolor{gray}{95.7} & \textcolor{gray}{94.8} & \textcolor{gray}{95.3} & \textcolor{gray}{95.2} & \textcolor{gray}{95.1} & & & & \\
\bottomrule
\end{tabular}
\end{table}

\textit{Training Efficiency} We conduct a comparative analysis of time cost across different methods and models. As shown in Tab.  \ref{tab:model_efficiency}, \sysname{} shows superior performance in less than 1.5 hours, far negligible than training or finetuning LLMs \cite{Thoppilan2022LaMDALM}.

\begin{table}[htbp]
\centering
\footnotesize
\caption{Training Efficiency Comparison}
\label{tab:model_efficiency}
\setlength{\tabcolsep}{3.5pt}
\begin{tabular}{@{}l cc ccc@{}}
\toprule
\multirow{2}{*}{Model} & \multicolumn{2}{c}{Vanilla Training} & \multicolumn{3}{c}{\sysname{} Training} \\
\cmidrule(lr){2-3} \cmidrule(lr){4-6}
& Time & Hardware & Time & Acc(\%) & Hardware \\
\midrule
BERT & 4 days & TPU×16 & 17min & 89.0 & A100×1 \\
GPT-Neo-2.7B & 90 days & A100×96 & 58min & 95.0 & A100×1 \\
GPT-J-6B & 5 weeks & TPU×256 & 89min & 95.0 & A100×1 \\
\bottomrule
\end{tabular}
\end{table}

\subsection{Extended Discussion on Limitations}
\label{sec:extend}

Our main experiments on Books3 and the Pile cover only a narrow slice of potential copyright data. Progress on dataset-level contribution analysis is further constrained by closed-source models, the scarcity of licensed copyright datasets, and unsettled legal laws. To mitigate these constraints, we complement the main experiments with a Matrix extension that expands coverage and improves transparency by pairing Matrix with the fully open-source LLM trained on it. 

\paragraph{Extended experiments on Matrix Dataset} We evaluate six Matrix sub-datasets: Book-Education, Book-Finance, Book-History, Book-Math, Book-Medical, and Book-Law \cite{zhang2024map}. These multilingual datasets better reflect broader and real-world attribution scenarios. Varying data scale, \sysname{} attains at least 89\% attribution accuracy and improves as the scale grows (Tab. \ref{tab:extend}). However, this means that for a specific dataset class containing extremely few samples, the performance degrades. In this case, a dedicated post-hoc detector can check outputs case by case \cite{mitchell2023detectgpt, inan2023llama}, since these rare samples may have limited influence on model behavior.

\paragraph{Effect of generation length} Contribution analysis accuracy slightly degrades when processing longer LLM outputs. In the continuation scene (as in Fig. \ref{fig:scenes}(a)), overall accuracy is always around 75\% but drops by about 3\% on average as length increases from 256 to 512 (Tab. \ref{tab:varylength}). Therefore, we recommend segmenting long LLM outputs into shorter ones before applying \sysname{}. In all, \sysname{} is accurate and scales with data on Matrix, but it weakens on very sparse classes and longer outputs; post-hoc checks and output segmentation are practical remedies.

\textbf{Scope and outlook}  Even with the Matrix extension, important gaps remain. Closed-source models and non-public copyright datasets still limit end-to-end attribution, and standardized, licensed benchmarks are still emerging. Broader access to them will enable more comprehensive evaluations of contribution analysis methods.

\begin{table}[H]
\centering
\footnotesize

\setlength{\tabcolsep}{4pt} 
\caption{ Extended Experiments on Map-Neo and Matrix Dataset. $k$ is set to 5 to show optimal performance.}
\label{tab:extend}
\begin{tabular}{@{}lccccc@{}}
\toprule
\makecell[c]{Data scale\\$k=5$} & Inter. & Var. & A-Var. & Baseline1 & Baseline2 \\
\midrule
\makecell[l]{Small\\(120,4800)} 
& 95.4 & 92.3 & 89.2 & 12.8 & 78.2 \\
\midrule
\makecell[l]{Medium\\(1200,4800)} 
& 97.9 & 97.5 & 97.6 & 12.8 & 86.7 \\
\midrule
\makecell[l]{Large\\(9600,4800)} 
& 99.8 & 99.2 & 99.0 & 12.8 & 89.5 \\
\bottomrule
\end{tabular}
\end{table}

\begin{table}[H]
\centering

\caption{ Contribution Analysis Accuracy with Varied Generation Length on Continuation Scene.}
\label{tab:varylength}
\begin{tabular}{@{}ccccc@{}}
\toprule
\makecell[c]{\textbf{Generation Length}\\\textbf{Accuracy(\%)}} & \textbf{64} & \textbf{128} & \textbf{256} & \textbf{512} \\
\midrule
Inter. & 73.65 & 72.95 & 77.82 & 71.73 \\
A-Var. & 72.95 & 72.96 & 80.56 & 78.94 \\
Var. & 70.98 & 77.03 & 75.15 & 72.98 \\
\bottomrule
\end{tabular}
\end{table}

\section{Conclusion and Future Work}
We highlight the need to identify specific copyrighted sub-datasets' influence on LLM outputs. Our proposed framework \sysname{} looks into LLM architectures, and discovers that MHA outputs provide more effective information than prompt-based, distance-based, or text classification methods. Information extraction, copyright sub-datasets contribution analysis and non-copyright responses filter modules are designed. \sysname{} shows high accuracy and AUC  comparable to text classification baselines when tested on original copyrighted texts. We further validate various generation scenarios, including prompt-induced book copyright infringements, continuation, copyright transfer, copyright elimination, and text mixture. \sysname{} efficiently combines multi-modules for copyrighted data protection with minimal data, time, and GPU memory need, outperforming text classification, prompt-based and distance-based baselines by more than 15.02\% accuracy and 0.104 AUC increase. Future work will extend to different fields such as Large Multimodal Models (LMMs) to verify its effectiveness in broader scenes.

\bibliographystyle{IEEEtran}
\bibliography{custom.bib}

\end{document}